%% file: main.tex
\documentclass[sigplan,nonacm]{acmart}
\usepackage{pifont}
\usepackage{balance}
\usepackage{enumitem}
\usepackage{graphicx}
\usepackage{subfigure}
\usepackage{multirow}
\usepackage{tablefootnote}
\usepackage{threeparttable}
\usepackage[vlined,linesnumbered,ruled]{algorithm2e}
\SetArgSty{textnormal} 
\usepackage{color}
\usepackage[noend]{algpseudocode}
\usepackage{xspace}
\usepackage{fancyhdr}
\usepackage{multicol}
\usepackage{amsfonts}
\usepackage{ulem}
\usepackage{tikz}
\usepackage{balance} 
\usepackage{makecell}
\usepackage{booktabs}

\def\ie{\textit{i.e.}\xspace}
\def\eg{\textit{e.g.}\xspace}

\def\method{\textsf{MoEpic}\xspace}
\def\PreGatedMoE{\textsf{Pre-gated MoE}\xspace}
\def\MixtralOffloading{\textsf{Mixtral-offloading}\xspace}
\def\AdapMoE{\textsf{AdapMoE}\xspace}
\def\MoEInfinity{\textsf{MoE-Infinity}\xspace}

\newcommand{\algnote}[1]{\textcolor[RGB]{148,0,211}{#1}}

\newcommand*{\circled}[1]{\lower.8ex\hbox{\tikz\draw (0pt, 0pt)%
    circle (.47em) node {\makebox[0.4em][c]{\small #1}};}}

\begin{document}
 
\title{Accelerating Mixture-of-Expert Inference with Adaptive Expert Split Mechanism}
\author{Jiaming Yan, Jianchun Liu, Hongli Xu, Liusheng Huang}

\begin{abstract}
\input{content/abs}
\end{abstract} 

% \keywords{LLM serving, Speculative inference, Multi-node collaboration.}

\maketitle

\section{Introduction}\label{sec_introduction}
\input{content/intro.tex}

\section{Background and Motivation}\label{sec:prelim}
\input{content/motivation.tex}

\section{System Design}\label{sec:design}
\input{content/design.tex}

\section{Performance Evaluation}\label{sec:evaluation}
\input{content/evaluation.tex}

\section{Related Work}\label{sec:related}
\input{content/works.tex}

\vspace{-3mm}
\section{Conclusion}\label{sec:conclusion} 
\input{content/conclusion.tex}

\bibliographystyle{ACM-Reference-Format}
\bibliography{main}

\end{document}

%% file: content/abs.tex
Mixture-of-Experts (MoE) has emerged as a promising architecture for modern large language models (LLMs).
However, massive parameters impose heavy GPU memory (\ie, VRAM) demands, hindering the widespread adoption of MoE LLMs.
Offloading the expert parameters to CPU RAM offers an effective way to alleviate the VRAM requirements for MoE inference. Existing approaches typically cache a small subset of experts in VRAM and/or dynamically prefetch experts from RAM during inference, leading to significant degradation in inference speed due to the poor cache hit rate and substantial expert loading latency.

In this work, we propose \method, an efficient MoE inference system with a novel expert split mechanism.
Specifically, each expert is vertically divided into two segments: top and bottom.
\method caches the top segment of hot experts, so that more experts will be stored under the limited VRAM budget, thereby improving the cache hit rate.
During each layer’s inference, \method predicts and prefetches the activated experts for the next layer.
Since the top segments of cached experts are exempt from fetching, the loading time is reduced, which allows efficient transfer-computation overlap.
Nevertheless, the performance of \method critically depends on the cache configuration (\ie, each layer's VRAM budget and expert split ratio).
To this end, we propose a divide-and-conquer algorithm based on fixed-point iteration for adaptive cache configuration.
Extensive experiments on popular MoE LLMs demonstrate that \method can save about half of the GPU cost, while lowering the inference latency by about 37.51\%$\thicksim$65.73\% compared to the baselines.

%% file: content/intro.tex
The advent of large language models (LLMs) has ushered in a transformative era for artificial intelligence (AI), revolutionizing various domains including natural language processing \cite{vaswani2017attention}, computer vision \cite{yuan2021tokens}, and multimodal tasks \cite{wu2024netllm}. 
These advancements are largely driven by the dramatic growth of model capacity \cite{kaplan2020scaling}, which entails significantly increased computational overhead during inference.
In this context, Mixture-of-Experts (MoE) has emerged as a promising architecture for LLMs, where the parameters are sparsely activated per input token, enabling sub-linear computational scaling relative to the model capacity \cite{cai2024survey}.
Recently, the MoE-based LLMs, such as DeepSeek-R1 \cite{guo2025deepseek} and Qwen-3 \cite{qwen3}, have demonstrated exceptional performance across diverse tasks.

Despite the computational efficiency of MoE models, their practical deployment faces significant hurdles, particularly in privacy-sensitive scenarios. 
Concretely, organizations in highly regulated sectors such as healthcare, financial services, and government institutions often mandate on-premises LLM deployment to ensure strict data privacy.
However, the modern MoE models with enormous parameter sizes impose substantial GPU memory (\ie, VRAM) requirements for inference, making local deployment prohibitively expensive.
For example, DeepSeek-R1 \cite{guo2025deepseek}, with its 671B parameters, demands 16 NVIDIA A100 (80GB) GPUs for storage, amounting to 300K USD in hardware expenses approximately.
Such expensive costs present a significant barrier to the widespread adoption of MoE models.

Fortunately, expert offloading \cite{eliseev2023fast} offers an efficient solution that significantly reduces the GPU requirements for MoE inference.
For instance, expert offloading enables DeepSeek-R1 to perform inference using only a single A100 GPU.
Specifically, in MoE models (using DeepSeek-R1 as a representative case), expert parameters typically constitute the vast majority (97.45\%) of the total model parameters, yet only a small fraction (3.13\%) are utilized per token.
This property  facilitates the VRAM optimization by offloading experts to spacious RAM or SSD, while only loading the activated ones on-demand into VRAM via the PCI express (PCIe).
Nevertheless, the transfer of experts between CPU and GPU will incur non-negligible loading latency, significantly degrading the Quality of Service (QoS), such as response speed.
For example, existing literature \cite{tang2024hobbit} demonstrate that the expert loading operation will slow down the inference speed by up to  17.18 times approximately.

To mitigate the impact of loading latency on the inference speed, researchers have explored various optimization approaches, which can be divided into two categories: cache-based and prefetch-based.
Firstly, the cache-based approaches \cite{eliseev2023fast, xue2024moe, tang2024hobbit} adopt the LRU or LFU policies to cache a subset of experts for each layer in VRAM.
Thus, the loading latency can be eliminated when the cached experts are activated (\ie, cache hit).
However, the number of cacheable experts is constrained by the limited VRAM budget, leading to a poor expert hit rate.
Empirical results from \textsf{ProMoE} \cite{song2024promoe} reveal that the cache hit rate will easily drop below 50\% under the constrained VRAM budget.
Consequently, over half of the model layers must load the activated experts on demand, which significantly prolongs inference latency.
Secondly, the prefetch-based approaches \cite{hwang2024pre, zhong2024adapmoe, song2024promoe} dynamically predict the experts needed for the next layer by analyzing the real-time intermediate result (\eg, activation value) during each layer's execution.
The anticipated experts can be prefetched proactively in parallel with the GPU computation to hide the loading latency.
Although these approaches can achieve considerable prediction accuracy (\eg, >80\% in \textsf{AdapMoE} \cite{zhong2024adapmoe}), the loading latency often exceeds the computation time (\eg,  165ms vs. 57ms according to our experiments in \S \ref{sec:motivation}).
Thus, only a small portion of loading latency can be overlapped, especially when handling multiple experts (\eg, DeepSeek-R1 activates 8 experts per token in each layer).

To conquer the above disadvantages, we expect to improve the cache hit rate under the limited VRAM budget, which severely restricts the number of cached (full-size) experts.
A natural solution is to only cache a part of each expert in VRAM.
To this end, we propose \method, an efficient \textbf{\underline{MoE}} inference system with a novel ex\textbf{\underline{p}}ert spl\textbf{\underline{i}}t me\textbf{\underline{c}}hanism.
Specifically, \method splits each expert into two segments, \ie, top and bottom, bringing two main advantages.
Firstly, \method selectively retains the top segments of some hot experts with high activation probabilities in VRAM.
Thus, under the same VRAM budget, \method significantly increases the number of cacheable experts compared to existing cache-based approaches that store full-size experts \cite{eliseev2023fast, xue2024moe, tang2024hobbit}.
% Secondly, during inference, \method utilizes the intemediate results  to predict and prefetch the most probable experts for the next layer.
% In this way, \method can achieve superior computation-transfer overlap compared to the existing works by balancing the computation time and loading latency.
% On the one hand, the top segments of cache-hit experts can be processed ahead of time while the missing experts (if any) are being fetched, extending the computation window.
% On the other hand, when the predicted/activated experts are already cached, only their bottom segments need to be fetched, which reduces the loading latency.
Secondly, during inference, \method utilizes the activation value of the current layer to predict the most probable experts for the next layer, and then prefetches their bottom segments (or full experts if missed).
In this way, the transfer-computation overlap can be optimized along two dimensions.
On the one hand, the top segments of cache-hit experts for the next layer will be processed ahead of time, extending the computation window (\ie, the time to overlap loading latency per layer).
On the other hand, the top segments of some cached experts will not be fetched, which reduces the loading latency.
By balancing the computation time and loading latency, this design enables more effective computation–transfer overlap than traditional prefetch-based approaches \cite{hwang2024pre, song2024promoe, zhong2024adapmoe}.

As a result, \method allows efficient deployment of MoE models over the resource-constrained devices while introducing negligible exposed loading latency.
However, the design of \method confronts two fundamental difficulties.
The first difficulty of \method lies in \textit{how to allocate the VRAM budget across different layers}.
Since the VRAM budget is shared across different layers, a natural strategy is to allocate it uniformly among all layers.
Nevertheless, according to our analysis in \S 2.3, the layers with low cache hit rates and/or poor expert prediction accuracy typically require more VRAM budget than others, making them potential bottlenecks under uniform allocation.
Conversely, allocating excessive budget to these layers would significantly prolong the inference latency of other layers, degrading the overall performance.
The second difficulty is \textit{How to split the expert properly for each layer} for \method.
Concretely, an oversized top segment prevents maintaining adequate experts within the limited VRAM budget, reducing the cache hit rates.
On the contrary, an excessively large bottom segment results in prolonged expert loading latency that cannot be effectively hidden by the computation time.
Besides, since the VRAM budget allocated to each layer varies, there is no universally optimal solution for all layers.
The main contributions of this paper are summarized as follows:
\begin{itemize}
    \item We propose an efficient MoE inference system with expert offloading, named \method. 
    By the novel expert split mechanism, \method can enhance cache hit rates while efficiently hiding the loading latency.
    \item To overcome the two challenges in \method's design, a divide-and-conquer algorithm based on fixed-point iteration is proposed for determining per-layer VRAM budget and split ratio.
    Besides, we introduce a priority-based cache policy for better cache management.
    \item Extensive experiments demonstrate that \method outperforms the state-of-the-art MoE inference systems with expert offloading.
    \method can save about half of the GPU cost, while lowering the inference latency by 37.51\%$\thicksim$65.73\% compared to the baselines.
\end{itemize}

%% file: content/motivation.tex
\begin{figure}[t]\centering
    \includegraphics[width=0.5\textwidth]{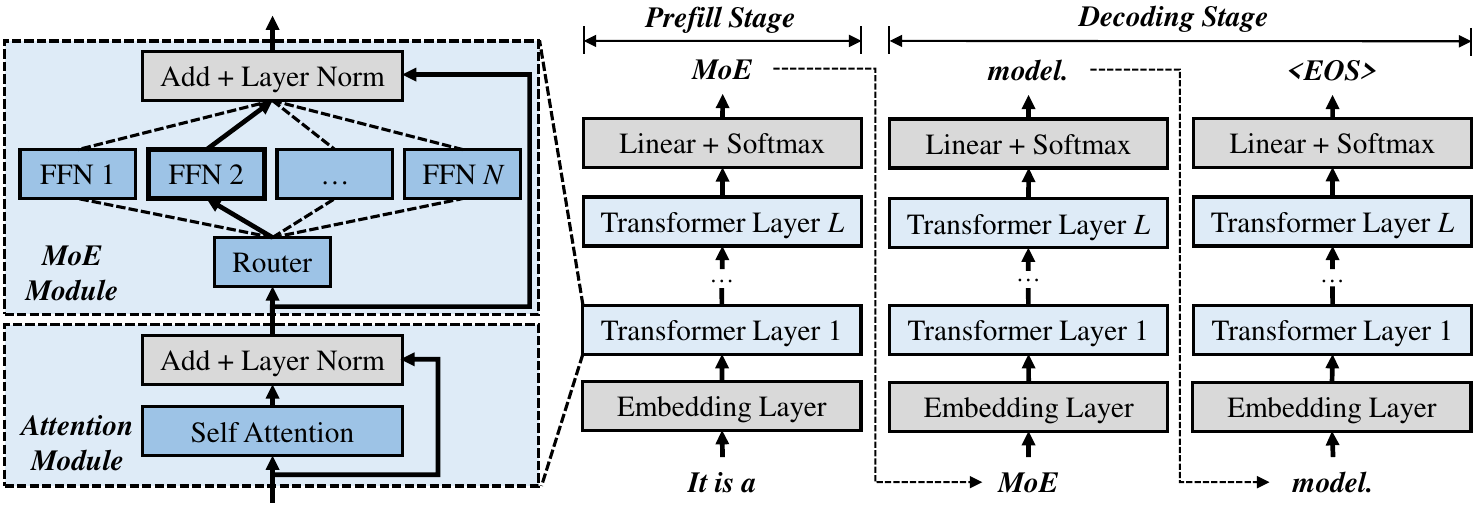}
    \caption{Inference process of the MoE model.} \label{fig:moe_inference}
    \vspace{-3mm}
 \end{figure}
 
\subsection{MoE Architecture \& Inference}

The architecture of the MoE-based LLM, as illustrated in Figure \ref{fig:moe_inference}, typically comprises multiple specialized Transformer layers stacked in sequence.
Specifically, considering an MoE model with $L$ layers, each layer $i \in [L]$ consists of two fundamental modules, \ie, an attention module followed by an MoE module.
Given an input token $x$, we use $x^{i}$ to denote the output of layer $i$, and $x^{0}$ to denote that of the embedding layer.
For each layer $i$, its input $x^{i - 1}$ (\ie, output of layer $i - 1$) is firstly processed by the attention module $\text{Att}^{i}$:
\begin{equation} \label{background_AttModule}
    h^{i} = \text{Norm}[x^{i - 1} + \text{Att}^{i}(x^{i - 1})] 
\end{equation}
where $\text{Norm} (\cdot)$ denotes the layer normalization operation \cite{xiong2020layer}, and $h^{i}$ represents the immediate result in layer $i$.
We refer readers to existing studies \cite{vaswani2017attention, ainslie2023gqa} for the detailed design of the attention module, as this is not our focus.

 \begin{figure*}[t]
	\centering
        \subfigure[Cache Performance.]{
		\includegraphics[width=1.65in]{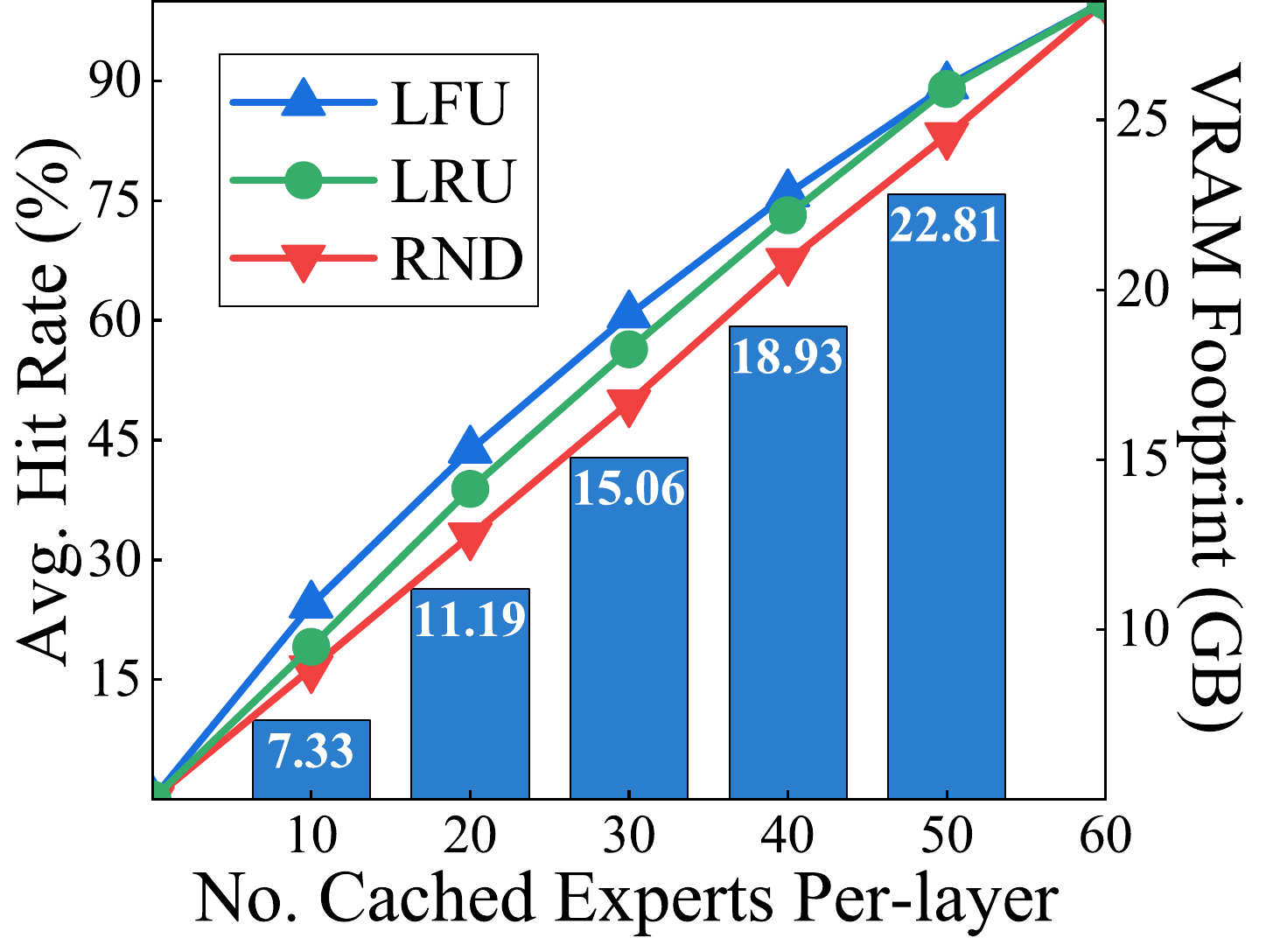}\label{fig:cache_performance}
	}
	\subfigure[Inference Latency.]{
		\includegraphics[width=1.65in]{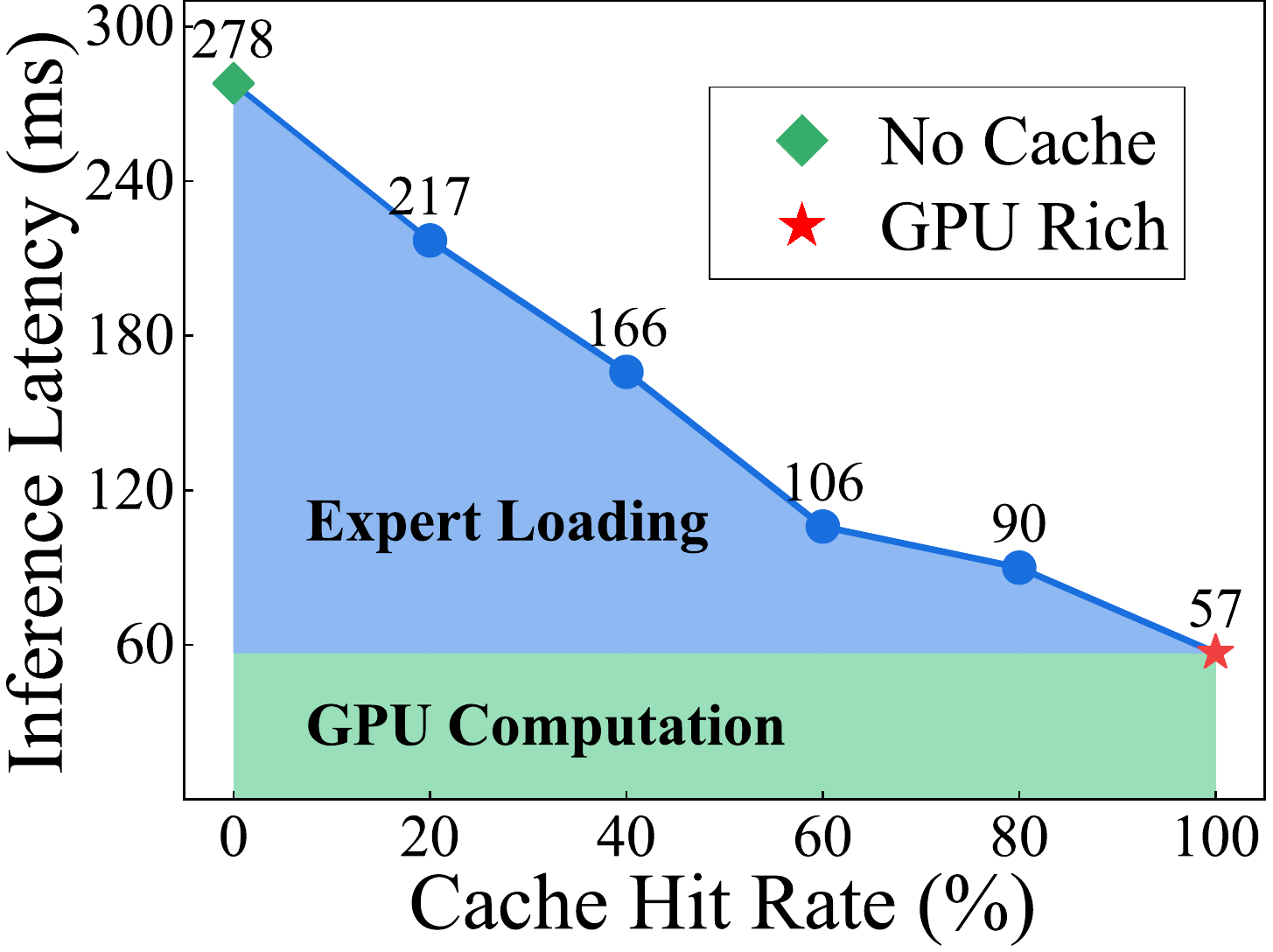}\label{fig:cache_latency}
	}
        \subfigure[Prediction Accuracy.]{
		\includegraphics[width=1.65in]{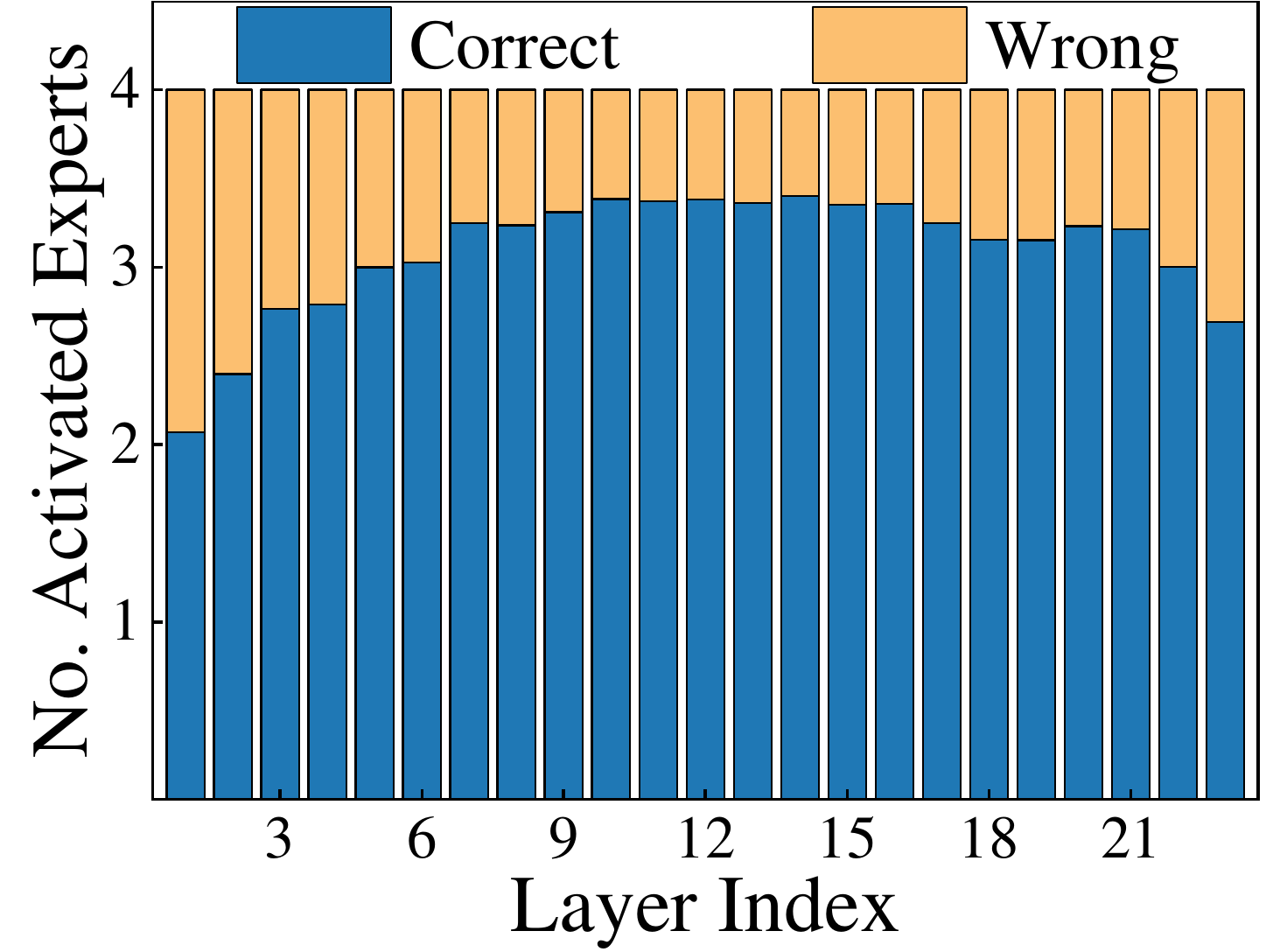}\label{fig:predict_accuracy}
	}
        \subfigure[Computation vs. load]{
		\includegraphics[width=1.65in]{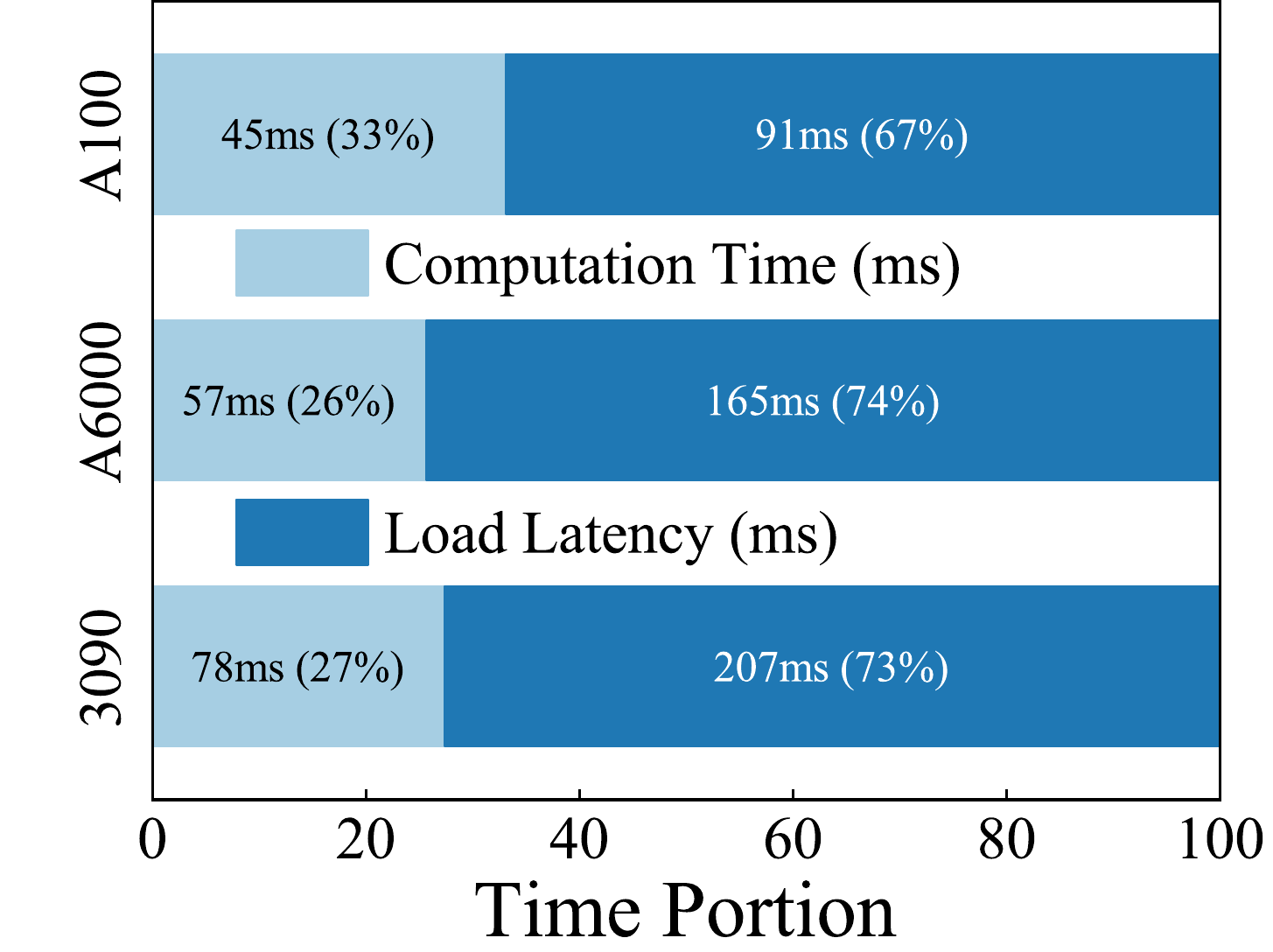}\label{fig:computation_communication}
	}
	\caption{
 The results of preliminary experiments using the Qwen1.5-MoE model on the MMLU dataset: 
(a) The VRAM footprint and average cache hit rate across all layers under different cache sizes (\ie, the number of cached experts per layer);
(b) The inference latency (\ie, time per output token) with given cache hit rates;
(c) The expert prediction accuracy of each layer;
(d) The per-layer computation time and the expert loading latency on different GPU platforms.
 }\label{fig:motivation}
 \vspace{-3mm}
\end{figure*}

\begin{figure}[t]\centering
    \includegraphics[width=0.5\textwidth]{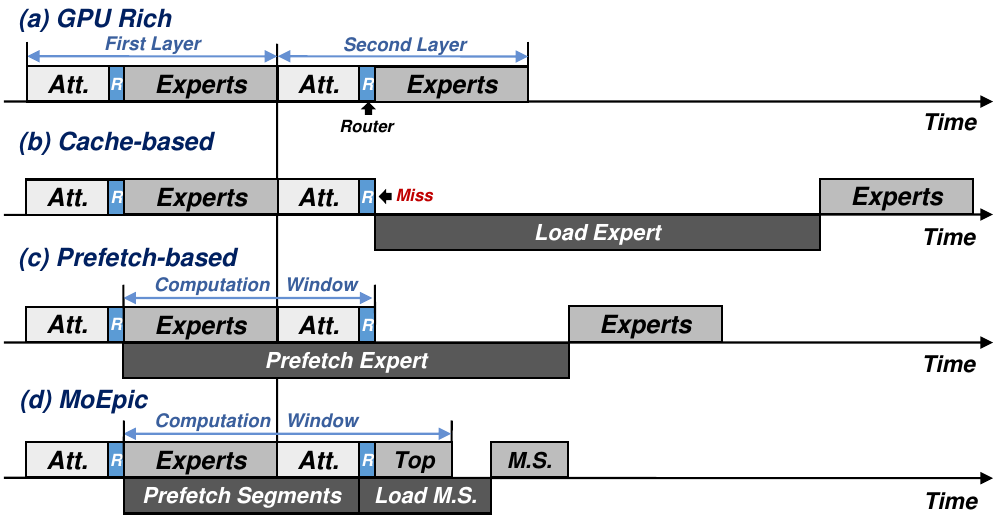}
    \caption{The inference process of two consecutive layers under different approaches. 
    For clarity of exposition, we do not include loading latency for the first layer, without loss of correctness.
    "Top" denotes the top segments of cache-hit experts. 
    "M.S." denotes the activated experts' missing (top/bottom) segments that are neither cached nor prefetched.} \label{fig:pipline_comparison}
    \vspace{-3mm}
 \end{figure}

Subsequently, $h^{i}$ is fed into the MoE module, which contains a router network $R^{i}$ ($\mathbb{R}^{d} \rightarrow \mathbb{R}^{N}$) and a set of feed-forward network (FFN) experts $\mathbf{E}^{i} = \{E_{i,1}, E_{i, 2}, \cdots, E_{i, N}\}$.
Here, $d$ represents the dimension of $h^{i}$, and $N$ denotes the total number of experts in each layer.
The router network calculates the score $s_{i, j} = \text{softmax}[R^{i}(h^{i})]_{j}$ of each expert $E_{i, j}$ ($j \in [N]$), then activates the $K$ ($< N$) experts with the highest scores.
Let $\mathcal{E}^{i}$ ($\subseteq \mathbf{E}^{i}$) represent the set of activated experts in layer $i$, and the output $x^{i}$ can be defined as:
\begin{equation} \label{background_MoeModule}
    x^{i} = \text{Norm}[h^{i} + \textstyle \sum_{E_{i, j} \in \mathcal{E}^{i}}  \frac{s_{i,j} \cdot E_{i, j} (h^{i})}{\sum_{E_{i, o} \in \mathcal{E}^{i}} s_{i, o}}]
\end{equation}

After the computation of all $L$ layers, the result $x^{L}$ is projected to a token probability distribution, from which the output token is sampled.
Same as the traditional LLMs (\eg, Llama-3 \cite{grattafiori2024llama}), the MoE inference follows an auto-regressive manner, which includes two stages: prefill and decoding.
The prefill stage processes all tokens in the input sequence simultaneously to generate the first token.
In the decoding stage, tokens are generated sequentially until the end-of-sequence token (\ie, <EOS>) is produced or the maximum output length is reached.
% Notably, during inference, the key and value vectors (\ie, KV cache)
% of processed tokens are retained to facilitate the computation of subsequent tokens.

\subsection{Limitations of Existing Approaches} \label{sec:motivation}
Due to the prohibitive cost of GPUs, the target devices (\eg, edge server) often lack sufficient VRAM capacity to host the entire MoE model.
To overcome this challenge, the most effective way is to offload the inactive experts to more capacious CPU RAM or SSD.
Existing approaches on expert offloading can be broadly classified into two main approaches: cache-based and prefetch-based.
For better clarity, we provide illustrative toy examples for the existing approaches and \method in Figure \ref{fig:pipline_comparison}.

\textbf{Cache-based Approaches.} This category of approaches focuses on storing only a small portion of experts in VRAM, avoiding loading latency when the activated experts are already in cache.
For example, \textsf{Mixtral-offloading} \cite{eliseev2023fast} supposes that consecutive tokens are likely to activate the same experts, thus it adopts the Least Recently Used (LRU) policy to manage the expert cache.
Differently, \textsf{MoE-Infinity} \cite{xue2024moe} caches the experts via the Least Frequently Used (LFU) policy.
However, these cache-based approaches often suffer from poor hit rates due to the limited number of cached experts.
To empirically validate such limitation, we conduct a set of preliminary experiments with the Qwen1.5-MoE model over the MMLU dataset \cite{hendrycks2020measuring}.
The tests are performed on an edge server equipped with 8 NVIDIA RTX A6000 GPU and PCIe 4.0. 
A more detailed experimental setup can be found in \S \ref{sec:setup}.
The results are presented in Figures \ref{fig:cache_performance}-\ref{fig:cache_latency}, where "RND" represents the method of randomly selecting cached experts.
As shown in Figure \ref{fig:cache_performance}, when caching 20 (out of 60) experts in each layer, the existing cache-based approaches achieve merely 33.02\%$\thicksim$43.74\% cache hit rates.
Accordingly, by Figure \ref{fig:cache_latency}, a cache hit rate of 40\% will introduce about 3.81$\times$ higher inference latency compared to the GPU-Rich baseline (\ie, the entire model fits within VRAM).
% In other word, VRAM budget limitations will lead to significant expert loading time for these cache-based approaches, as illustrated in Figure \ref{fig:pipline_comparison}(b).
In a word, with the  constraint of VRAM budget, the frequent cache misses will increase the inference latency significantly compared to the GPU Rich setting.

\textbf{Prefetch-based Approaches.}
These approaches try to overlap the loading latency with the computation time, to accelerate the MoE inference.
Specifically, during the computation of the current layer, the system predicts and prefetches the experts needed for the next layer in advance.
For instance, \textsf{Pre-gated MoE} \cite{hwang2024pre} modifies the model architecture by directly connecting each layer's attention module to the router network of its subsequent layer, while fine-tuning the router network for accurate expert prediction.
Moreover, building upon the observed high similarity of intermediate representations between consecutive layers, \textsf{AdapMoE} \cite{zhong2024adapmoe} utilizes the intermediate result of the current layer to feed the router network of the next layer to predict the required experts.
Figure \ref{fig:predict_accuracy} demonstrates that these methods maintain considerable prediction accuracy across different model layers.
That is, among the 4 experts activated in each layer of Qwen1.5-MoE, an average of 2.07$\thicksim$3.39 experts will be prefetched correctly.
With correct prefetching, this approach enables the expert loading latency to be overlapped with the computation of the previous layer’s MoE module and current layer’s attention module.
However, the potential for transfer-computation overlap remains under-exploited in current implementations.
Specifically, we measure the GPU computation time and the expert loading latency for processing each token across different GPU platforms (\ie, NVIDIA  RTX 3090, A6000, and A100), where the results are shown in Figure \ref{fig:computation_communication}.
We find that merely 34.55\%$\thicksim$49.45\% of the loading latency can be hidden.
By Figure \ref{fig:pipline_comparison}(c), the uncovered time directly adds to end-to-end inference latency.
Therefore, the imbalance between computation time and loading latency limites the efficiency of these approaches.

\subsection{Motivations for System Design}
In summary, the inference performance of existing approaches is primarily constrained by the poor cache hit rate and high expert loading latency.
% To address these two issues, both the number of cached experts and loading efficiency should be improved.
% Motivated by this objective, 
Motivated by these two issues, we present a novel MoE inference framework, named \method, which vertically splits each expert into two segments: top for static caching and bottom for dynamic prefetching.
This design enables caching more experts within limited VRAM budget, thereby improving the cache hit rates significantly.
As depicted in Figure \ref{fig:pipline_comparison}(d), the expert split mechanism also improves the loading efficiency.
Specifically, with the top segments cached in VRAM, only the bottom segments need to be fetched for hot experts, reducing loading latency.
Consequently, more experts can be prefetched within the per-layer computation time.
Besides, the top segments of cache-hit experts can be processed first, even before their bottom segments are loaded into VRAM, thereby extending the computation window to overlap the loading latency of missing segments.
However, to unleash the full potential of \method, we need to address the following two key difficulties:

\textbf{VRAM Budget Allocation.}
In practical deployments, naively allocating VRAM budget evenly across all layers will lead to sub-optimal performance for \method.
% We split each expert in half and cache the top segments of 30 experts per layer, using the LFU policy for cache management.
% We then record the per-layer inference latency, as shown in Figure \ref{fig:perlayer_latency}.
To validate this phenomenon, we measured per-layer inference latency with the LFU policy, where each expert is divided in half, and the top segments of 30 experts are cached for each layer. 
% Specifically, we split each expert in half and cached the top segments of 30 experts per layer, while employing the LFU policy for cache management.
It can be observed that several layers (\eg, layers 3, 10, and 21) exhibit significantly higher latency than others.
We attribute this to the fact that two types of layers require more VRAM budget to efficiently hide the loading latency.
1) According to our experiments, the cache hit rate varies by up to 5.55\% across different layers (58.12\% vs. 63.67\%).
For layers with relatively low cache hit rates, additional VRAM budget is necessary to cache more experts and maintain adequate hit rate.
2) For layers with low expert prediction accuracy, \eg, those located at the head and tail of the model, as shown in Figure \ref{fig:predict_accuracy}, increasing the budget enables larger top segments, which reduces the loading cost for the cached exeprts.
Thus, the expanded VRAM budget allows more experts to be prefetched within the computation window, improving the likelihood of correct prefetching.
However, assigning too much budget to these layers will severely impact the inference speed of other layers.
Consequently, \textit{how to properly allocate the VRAM budget across different layers} presents the first difficulty in \method.

 \begin{figure}[t]
	\centering
        \subfigure[Per-layer Latency.]{
		\includegraphics[width=1.58in]{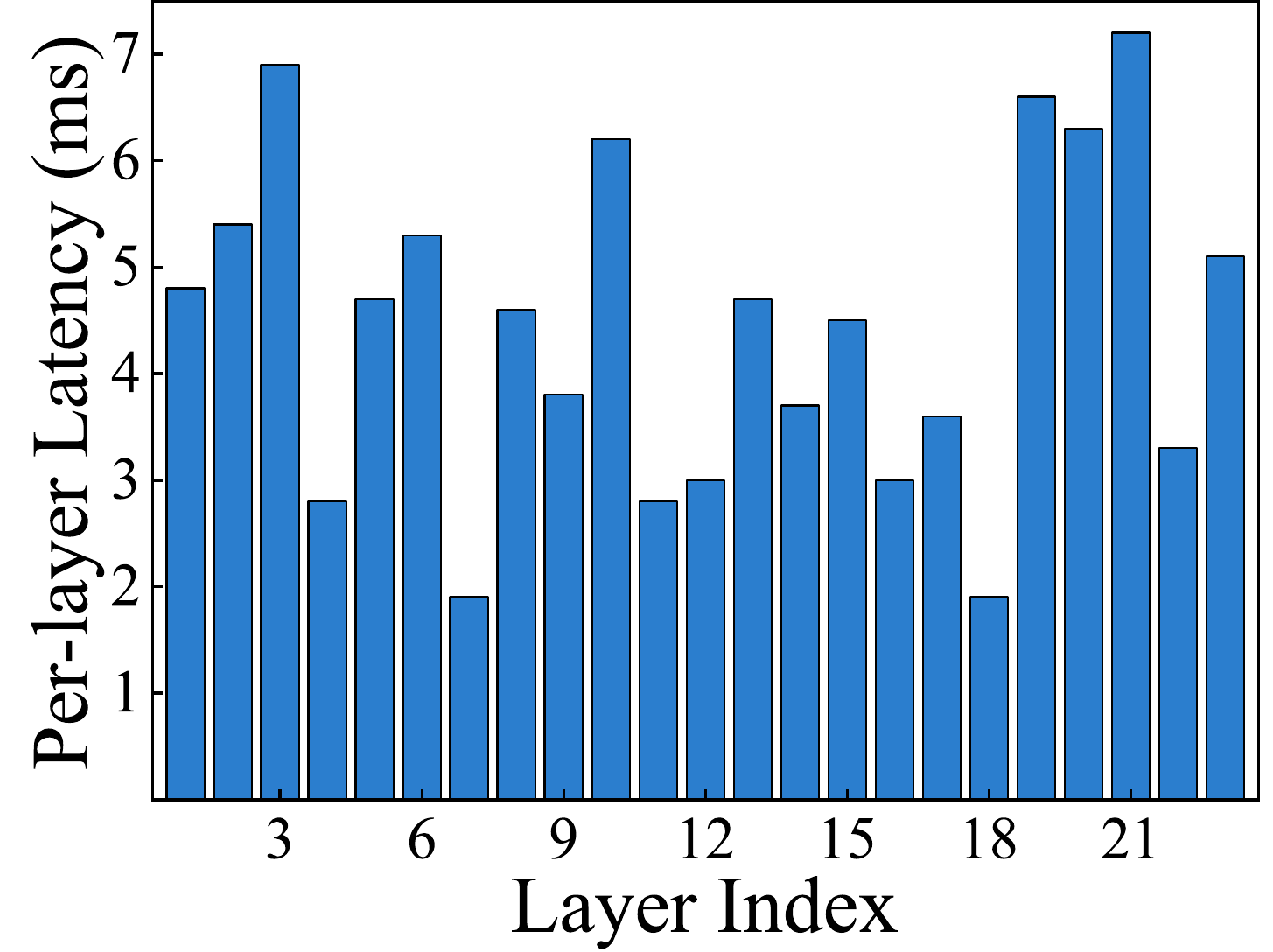}\label{fig:perlayer_latency}
	}
	\subfigure[Impact of Varied Split Ratios.]{
		\includegraphics[width=1.58in]{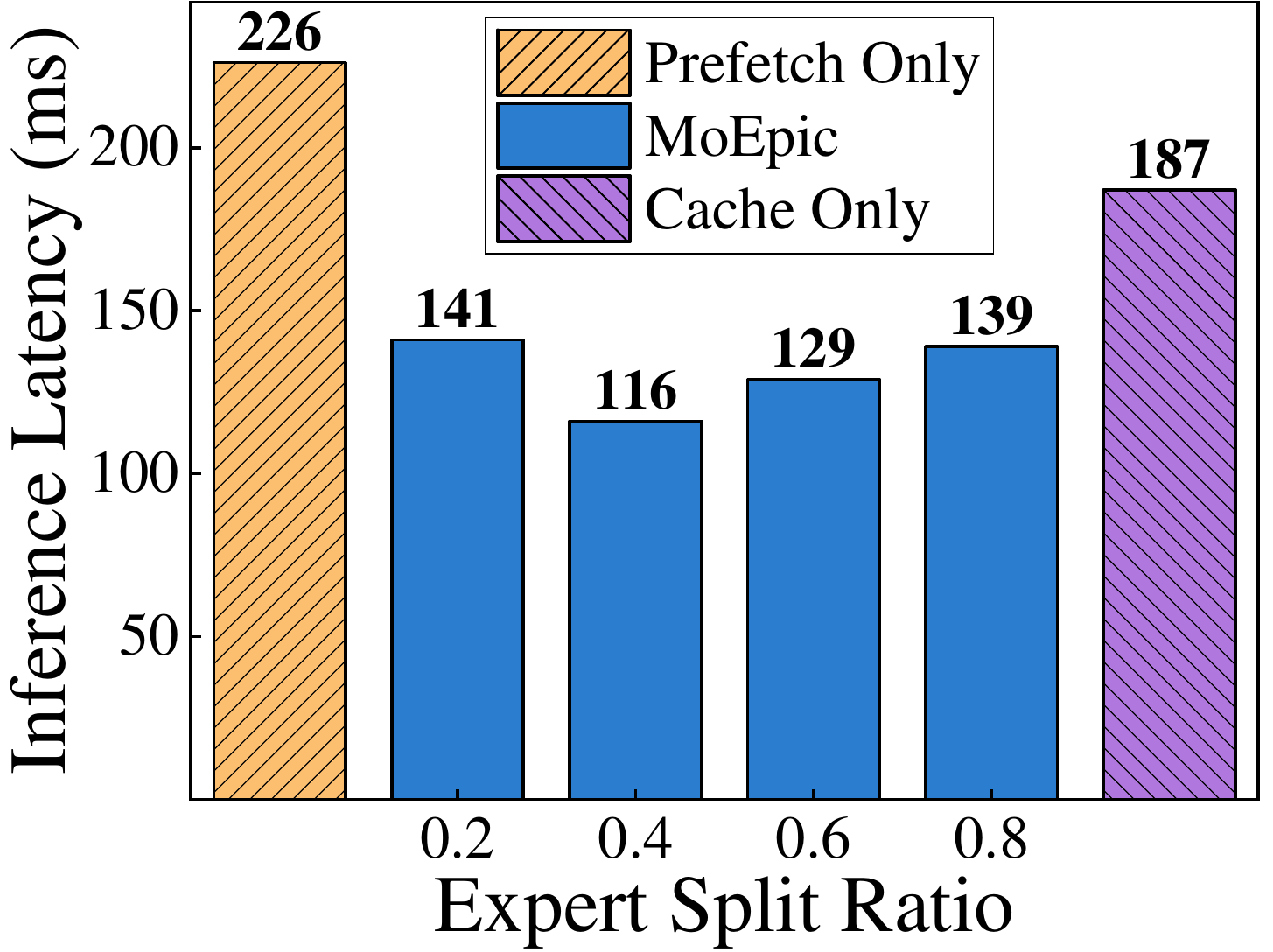}\label{fig:split_ratio_impact}
	}
    \vspace{-3mm}
	\caption{Motivations for \method's design. }\label{fig:insight}
    \vspace{-4mm}
\end{figure}

\textbf{Split Ratio Determination.}
According to our empirical observation, the performance of \method critically depends on the expert split ratio $\theta$, which is defined as the proportion of the top segment relative to the full expert.
It is worth noting that when $\theta = 0$, \method degenerates into the prefetch-based approach; when $\theta = 1$, and expert prefetching is disabled, \method degenerates into the cache-based approach.
Here, we include these two cases as baseline approaches for comparison, denoted as \textsf{Prefetch Only} and \textsf{Cache Only}, respectively.
As illustrated in Figure \ref{fig:split_ratio_impact}, under the same VRAM budget of 288 full-size experts (12 per layer),  varying split ratios exhibit substantial differences in inference latency, with the optimal performance achieved at $\theta$=0.4.
At this sweet spot, \method can reduce the inference latency by about 37.97\%$\thicksim$48.67\% compared to baselines, providing preliminary evidence for the effectiveness of the expert split mechanism.
The results also underscore the necessity to optimize the expert split ratio.
Nevertheless, identifying the optimal split ratio remains non-trivial, as it requires jointly considering the target device’s computation speed, PCIe bandwidth, and per-layer VRAM budget.
An overly small ratio may result in prolonged bottom segment loading latency, which cannot be effectively masked by computation time.
Conversely, an excessively large ratio will restrict the number of cacheable experts and consequently reduce the cache hit rate.
Thus, the second difficulty for \method is \textit{how to determine the optimial expert split ratio for each layer.}

%% file: content/design.tex
In this section, we first present the system overview of \method, followed by the descriptions of its three key components.
We begin with the speculative experts prefetcher.
Then, a new priority-based metric is introduced for better cache management.
Finally, we formulate the problem for optimizing the cache configuration, and design an efficient divide-and-conquer algorithm based on fixed-point iteration to solve the optimization problem.

\subsection{System Overview}
\begin{figure}[t]\centering
    \includegraphics[width=0.45\textwidth]{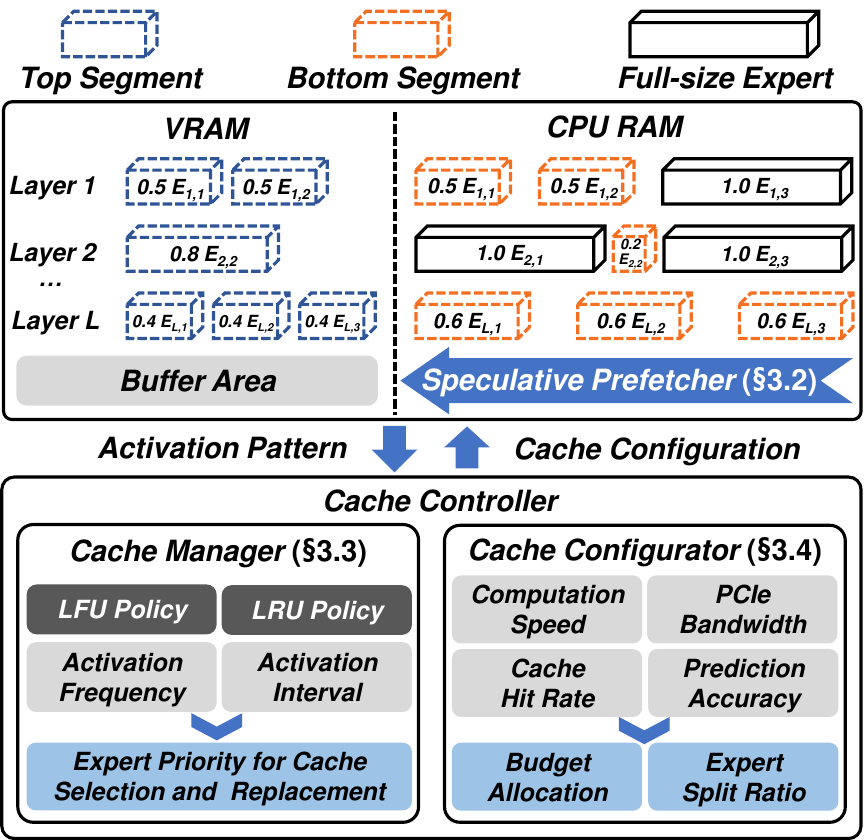}
    \caption{System overview of \method.} \label{fig:system_overview}

 \end{figure}

We provide a system overview of \method with a toy example in Figure \ref{fig:system_overview}.
During the inference of each layer, \method employs a \textit{Speculative Prefetcher} (detailed in \S \ref{sec:prefetch})  to proactively load the bottom segments (or full-size experts) for the next layer into VRAM's buffer.
The prefetching executes concurrently with the GPU computation, so that the expert loading latency can be hidden.
If the cache hit and the prefetching is correct, the cached top segment and the buffered bottom segment are concatenated to perform the MoE module computation.
Otherwise, \method loads the missing segments of the activated experts while concurrently processing available segments in VRAM.
Additionally, \textit{Cache Controller} analyzes the expert's activation pattern in each layer and optimizes the cache management and configuration.
Specifically, it consists of two key components.
Firstly, \textit{Cache Manager} (detailed in \S \ref{sec:lcp}) computes a priority score for each expert according to both its activation frequency and activation interval, guiding the cache selection and replacement.
This strategy combines the advantages of LFU and LRU, improving the cache hit rate under the given cache size.
Secondly, \textit{Cache Configurator} (detailed in \S \ref{sec:algorithm}) considers the hardware specifications (\ie, computation speed, PCIe bandwidth) of target device, as well as the current performance metrics (\ie, cache hit rate, expert prediction accuracy) to jointly determine the VRAM budgets and split ratios for different layers.
When the device is idle, \method adjusts the expert cache in VRAM based on the updated cache configuration for better performance.
For instance, as shown in Figure \ref{fig:system_overview}, layer 1 has a cache size of 2 and a split ratio of 0.5, whereas layer $L$ has a cache size of 3 and a split ratio of 0.4.
For layer 2, $E_{2, 2}$ is the expert with the highest cache priority.

\subsection{Speculative Prefetcher} \label{sec:prefetch}
\begin{figure}[t]
	\centering
        \subfigure[Qwen1.5-MoE.]{
		\includegraphics[width=1.58in]{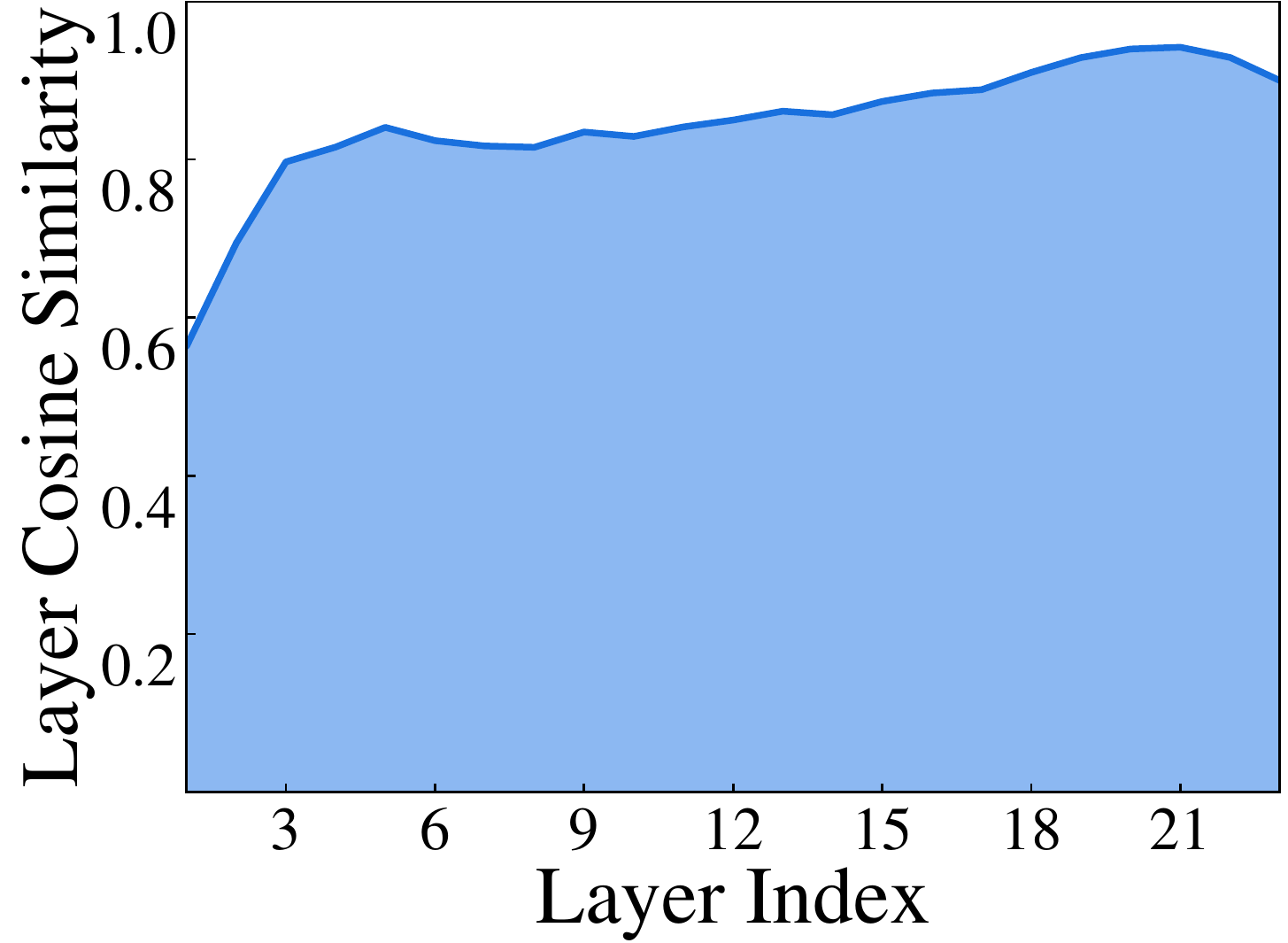}\label{fig:qwen_similarity}
	}
	\subfigure[Mixtral-8x7B.]{
		\includegraphics[width=1.58in]{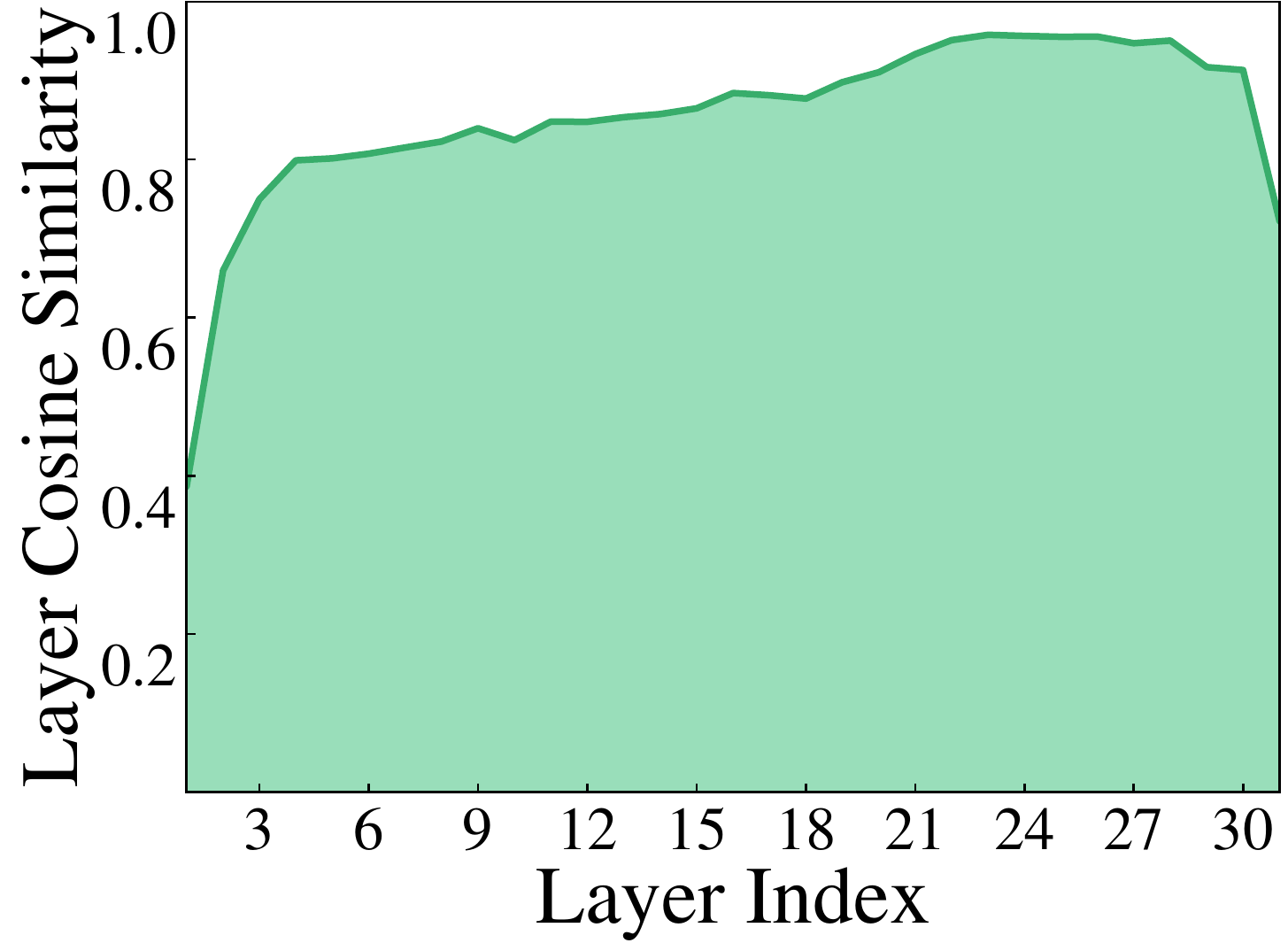}\label{fig:mixtral_similarity}
	}
    \vspace{-3mm}
	\caption{Layer-wise cosine similarity of two models.}\label{fig:similarity}
\end{figure}

With the advantages in training performance (\eg, gradient stability, convergence rate), residual connection \cite{he2016deep} has become a fundamental architecture in modern LLMs.
Due to the nature of residual structure, the intermediate activation (\ie, attention module's output $h^{i}$) between two successive layers often exhibits remarkable similarity \cite{zhong2024adapmoe}.
To empirically validate this phenomenon, we measure the cosine similarity between $h^{i}$ and $h^{i +1}$ with two MoE models, where a similarity value of 1 indicates identical representations.
As shown in Figure \ref{fig:similarity}, both these two models demonstrate similar intermediate activation across adjacent layers.
It reveals that the intermediate activation of each layer can be speculatively approximated as that of its previous layer.
Thus, during the forward pass of each layer $i$, \method feeds the intermediate activation $h^{i}$ to the next layer's router network $R^{i + 1}$, for predicting the score of each expert $j$ in layer $i + 1$:
\begin{equation}
    s_{i + 1, j}^{pred} = \text{softmax} [R^{i + 1}(h^{i})]_{j}, \quad i \in [L - 1], j \in [N]
\end{equation}
Upon obtaining the results from the router network at each layer $i$, \method terminates the prefetch operation and checks whether all activated experts have been fully loaded into VRAM.
To mitigate the GPU idle time, \method then simultaneously performs two operations: 1) initiates loading the missing experts and/or bottom segments (if have) from RAM, while 2) processes experts and/or top segments that are already residing in VRAM, overlapping GPU computation and experts loading.
After fully loading all activated experts at layer $i$, \method prefetches the experts for the next layer $i+1$ to the VRAM buffer.
Specifically, the experts are prefetched one by one in descending order of their predicted scores.
Notably, since the first layer lacks previous results, we follow \AdapMoE \cite{zhong2024adapmoe} to use the last layer's intermediate activation of previous token for expert prediction.
For each target expert, \method prefetches only the bottom segment if its top segment is cached; otherwise, the full expert is required to be prefetched.
As shown in Figure \ref{fig:pipline_comparison}(d), the prefetch operation for layer $i + 1$ is parallel with the computation of layer $i$'s MoE module and layer $i + 1$'s attention module.
However, there exists a potential issue for this speculative prefetch method. 
By Figure \ref{fig:similarity}, the intermediate activation shows relatively low similarity across the initial layers, leading to poor expert prediction accuracy, which is also validated in Figure \ref{fig:predict_accuracy}.
As a result, the initial layers may need more VRAM than other layers, allowing caching more experts and/or adopting larger split ratios.
Thus, more experts can be prefetched within the computation window, mitigating the incidence of expert misses significantly.
These analysis motivates our design in \S \ref{sec:algorithm}.

\subsection{Priority-based Cache Manager} \label{sec:lcp}

Building upon prior findings \cite{eliseev2023fast, zhuang2024litemoe}, the expert activation patterns demonstrate two fundamental properties in most layers.
(1) Long-tail distribution: some "hot" experts are much more frequently activated than others \cite{zhuang2024litemoe}.
(2) Temporal locality: consecutive tokens often show a preference for activating the same expert \cite{eliseev2023fast}.
It is evident that the LFU policy is well-suited for capturing the long-tail distribution, whereas the LRU policy excels at handling the temporal locality.
However, neither LRU nor LFU adequately considers both properties simultaneously, leading to sub-optimal cache hit rates.
To address this issue, we introduce a new metric, named \textit{cache priority}, to enable better cache management.

Specifically, \method records two key indicators for every expert $E_{i, j}$ at layer $i$ during inference.
The first is activation frequency $\mu_{i, j}$, which is defined as the accumulated activation count of the expert.
The second is activation interval $\nu_{i, j}$, tracking how many consecutive tokens have not activate this expert at present.
Based on these two indicators, each expert $E_{i, j}$'s cache priority $\mathbb{P}_{i, j}$ can be computed as:
\begin{equation} \label{cache_priority}
    \mathbb{P}_{i, j} = \mu_{i, j} \cdot \rho^{\nu_{i, j} / \omega}
\end{equation}
where $\omega$ is the size of observation window. 
$\rho \in (0, 1)$ is a hyper-parameter that controls the importance of $\mu_{i, j}$ and $\nu_{i, j}$.
In our experiments, $\omega$ and $\rho$ are separately set to 128 and 0.25 by default.
Once the cache miss, \method evicts the lowest-priority expert from VRAM, and replaces it with the newly loaded expert.
We name this priority-based policy as LCP (Least Cache Priority).
Clearly, LCP models both the long-tail distribution and temporal locality properties inherent in the expert activation patterns, effectively combining the benefits of both the LRU and LFU policies.

We conducted a set of preliminary experiments to validate the performance of LCP, with the results presented in Table \ref{table:policy_comparison}.
It can be found that LCP can effectively improve the cache hit rate.
For instance, when caching 30 experts per layer, LCP achieves 12.52\%, 1.53\% and 5.83\% higher hit rate compared to RND, LRU and LFU, respectively.
Despite LCP yields improvements compared to the baselines, the cache hit rate remains fundamentally constrained by the cache size, which motivates \method's design in scaling the number of cacheable experts under the limited VRAM budget.

\begin{table}[t]
\centering
\caption{Comparison of cache hit rates across different policies with the Qwen1.5-MoE model and the MMLU dataset.}
\begin{tabular}{c|ccccc}
\toprule
\multirow{2}{*}{Policy} & \multicolumn{5}{c}{Number of Cached Experts Per Layer} \\
& 10 & 20 & 30 & 40 & 50   \\
\midrule
 RND     & 16.41\% & 33.02\% & 49.69\% & 67.41\% & 83.15\% \\
 LFU     & 24.36\% & 43.74\% & 60.68\% & 75.84\% & 89.34\% \\
 LRU     & 19.11\% & 38.88\% & 56.38\% & 73.16\% & 88.97\% \\
 LCP     & \textbf{25.56\%} & \textbf{45.36\%} & \textbf{62.21\%} & \textbf{77.12\%} & \textbf{90.08\%} \\
\bottomrule
\end{tabular}
\label{table:policy_comparison}
\end{table}

\subsection{Cache Configurator} \label{sec:algorithm}

\subsubsection{Problem Formulation}

We consider a MoE model $\mathbf{M}$ consists of $L$ layers, each with $N$ experts, where only $K$ ($\ll N$) experts are sparsely activated per token.
Given the target device $\mathbf{D}$, we denote its PCIe bandwidth and VRAM capacity as $B$ and $V$, respectively.
Let $U$ be the total model size, with $U_{n}$ indicating the size of all non-expert parameters.
Hence, the size of each expert can be expressed as $U_{e} = \frac{U-U_{n}}{LN}$, and the time for loading an expert is given by $T_{load}^e = U_{e}/B$.
We can profile each module's execution time of model $\mathbf{M}$ over device $\mathbf{D}$, including that of attention module $T_{comp}^{att}$, MoE module $T_{comp}^{moe}$, and head layer $T_{comp}^{head}$.
We omit the computation time of router network, as it constitutes a negligible fraction of the overall inference latency.
Besides, $T_{wind}^{i}$ is defined as the available time window for prefetching experts in layer $i$.
For the first layer, the prefetching window involves the time for processing the previous token in the head layer and the time for the attention module in the first layer, \ie, $T_{wind}^{1} = T_{comp}^{head} + T_{comp}^{att}$.
During each prefetching window $T_{wind}^{i}$, \method proactively prefetches experts for layer $i$ by descending order of their predicted scores.

Let $V_{i}$ and $C_{i}$ represent the allocated VRAM budget and the cache size (\ie, number of cached experts) in layer $i$, respectively.
The expert split ratio is then computed as $\theta_{i} = V_{i} / C_{i}$.
Given the cache size $C_{i}$ and prefetching window $T_{wind}^{i}$ of each layer $i$, we assume that after the process of layer $i$'s router network, $\alpha_{i}$ activated experts are fully loaded into VRAM, $\beta_{i}$ activated experts have only their top segments cached in VRAM, and $\gamma_{i}$ activated experts remain completely offloaded in RAM.
The values of $\alpha_{i}$, $\beta_{i}$, and $\gamma_{i}$  are critically dependent on both the cache hit rate and the expert prediction accuracy of layer $i$.
As previously introduced, once obtaining the routing result, \method starts to process the already-loaded activated experts while simultaneously fetching the missed experts.
Let $T_{comp}^{exp} = T_{comp}^{moe} / K$ represents the time for processing one expert.
Consequently, we can formulate the expert computation time for hiding the missed expert loading latency as follows:
\begin{equation} \label{miss_window}
    T_{hide}^{i} = (\alpha_{i} + \beta_{i} \theta_{i} ) \cdot T_{comp}^{exp}, \quad i \in [L]
\end{equation}

We denote the latency for fetching an expert from RAM to VRAM as $T_{load}^{exp}$.
Then, the latency for loading the missed experts and/or segments in layer $i$ is given by $T_{miss}^{i} = [\beta_{i} (1 - \theta_{i}) + \gamma_{i}] \cdot T_{load}^{exp}$.
Finally, the expert loading latency that can not be overlapped by the GPU computation is formulated as:
\begin{equation} \label{non_overlapped_loading_latency}
    \mathcal{T}^{i} = \max \{0, T_{miss}^{i} - T_{hide}^{i} \}, \quad i \in [L]
\end{equation}

At this point, we can derive the expected value of each layer $i$'s prefetching window $T_{wind}^{i}$ ($i > 1$):
\begin{equation} \label{prefetch_window}
    T_{wind}^{i} =  (T_{comp}^{moe} - \min \{ T_{hide}^{i - 1}, T_{miss}^{i - 1} \}) + T_{comp}^{att}, \quad 2 \leq i \leq L
\end{equation}

To improve the response speed of the MoE inference with expert offloading, the optimization problem in \method aims to minimize the non-overlapped expert loading latency by determining the optimal VRAM budgets and expert split ratios across different layers:
$$ \min_{ V_{i}, \theta_{i} (i \in [L])} \sum_{i = 1}^{L} \mathcal{T}^{i} $$ 
\begin{equation} \label{problem_formultation}
    s.t.
    \begin{cases}
        U_{n} + \sum_{i = 1}^{L} V_{i} + U_{b} \leq V & \\
        \alpha_{i} + \beta_{i} + \gamma_{i} = K,      & \forall i \in [L] \\
        V_{i} > 0, \theta_{i} \in (0,1],              & \forall i \in [L] 
    \end{cases}
\end{equation}
where the first inequality meets the VRAM constraint, and $U_{b}$ indicates the buffer size.
The second set of inequalities indicates that the activated experts in each layer can only be in three statuses: fully loaded, only the top segment loaded, or completely unloaded.
The third sets of inequalities specify the feasible ranges for the optimization variables $V_{i}$ and $\theta_{i}$ ($i \in [L]$), respectively.

\subsubsection{Algorithm Design}

However, since obtaining the precise values of $\alpha_{i}$, $\beta_{i}$, and $\gamma_{i}$ for each layer $i$ is quite difficult in practice, it is infeasible to directly solve this optimization problem.
To tackle this issue, we design a divide-and-conquer algorithm based on fixed-point iteration to optimize the cache configuration of \method.

To begin with, we decompose the complex and intractable optimization problem in Eq. \eqref{problem_formultation} and focus on a more manageable sub-problem.
Specifically, given the VRAM budget $V_{i}$ and the prefetching window $T_{wind}^{i}$ for a specific layer $i$, we aim to determine the optimal expert split ratio $\theta_{i}$ that minimizes the exposed loading latency $\mathcal{T}^{i}$.
To solve this sub-problem, \method records the (predicted) experts scores for each token at every layer, and derives the following three probabilities for layer $i$ after the inference of $q$ tokens:
\begin{enumerate} [label=\arabic*)]
    \item Cache Hit Rate $\mathbf{H}_{i} (C_{i})$ is defined as the number of times the activated experts hit the cache with a cache size of $C_{i}$ ($\in [N]$), divided by the total number of activated experts (\ie, $q \cdot K$).
    \item Expert Prediction Accuracy $\mathbf{P}_{i}(y)$ is defined as the activation frequency of the expert with $y$-th highest predicted score ($y \in [N]$), divided by the total number of processed tokens (\ie, $q$).
    \item Prediction Cache Hit Rate $\mathbf{PH}_{i} (y, C_{i})$ is defined as the number of times the expert with $y$-th highest predicted score hits the cache with a cache size of $C_{i}$, divided by the total number of processed tokens.
\end{enumerate}

Then, we relax the discrete count of activated experts into a continuous optimization variable.
Concretely, we assume that after the prefetch operation with window size $T_{wind}^{i}$, $m_{i}$ ($\in [0, K]$) activated experts are already loaded in VRAM.
At this point, the goal of minimizing exposed loading latency $\mathcal{T}^{i}$ is equivalent to maximizing $m_{i}$.
Given a cache size of $C_{i}$, the amount of cached activated experts is $A_{cache}^{i} (\theta_{i}, C_{i}) = K \cdot \mathbf{H}_{i} (C_{i}) \cdot \theta_{i}$.
Consequently, considering the prefetch operation for the $y$-th predicted expert (\ie, expert with $y$-th highest predicted score), its expected execution time $T_{pref}^{i} (a, C_{i})$ can be computed as follows:
\begin{equation} \label{prefetch_k_time}
     T_{pref}^{i} (y, C_{i}) = [1 -  \mathbf{PH_{i}} (y, C_{i}) \cdot  \theta_{i}] \cdot T_{load}^{exp}
\end{equation}

Moreover, we can quantify the expected loading amount of activated experts $A_{pref}^{i} (y, C_{i})$ under this prefetch operation:

\begin{equation} \label{prefetch_k_amount}
     A_{pref}^{i} (y, C_{i}) =  [1 -  \mathbf{PH_{i}} (y, C_{i}) \cdot  \theta_{i}] \cdot \mathbf{P}_{i} (y, C_{i})
\end{equation} 

Suppose that at most $Y$ expert prefetch operations can be completed within the window $T_{wind}^{i}$.
Here, the formulation for the sub-problem optimization is given as the following Eq. \eqref{subproblem_formultation}, which can be easily solved by searching the optimal cache size $C_{i}^{*}$.
\vspace{-3mm}
$$ \max_{ C_{i} \in [N] }  m_{i} = A_{cache}^{i} (\theta_{i}, C_{i}) + \sum_{y = 1}^{Y} A_{pref}^{i} (y, C_{i}) $$ 
\begin{equation} \label{subproblem_formultation}
    s.t.
    \begin{cases}
        \theta_{i} = V_{i} / C_{i}                  \\
        \sum_{y = 1}^{Y} T_{pref}^{i}  (y, C_{i}) \leq T_{wind}^{i}
    \end{cases}
\end{equation}

\begin{algorithm}[t] 
\caption{Cache Configuration Algorithm} \label{alg}
\KwIn{Latency metrics $T_{comp}^{att}$, $T_{comp}^{moe}$, $T_{comp}^{head}$, and $T_{load}^{exp}$, VRAM budget for caching experts $V_{e}$ ($=V - U_{n} - U_{b}$), cache update frequency $\tau$, allocation granularity $\zeta$
}
Set $q$ to 0 \quad \algnote{ \# the number of processed tokens} \\ \label{alg:init_begin}
Set $\mathbb{P}_{i, j}$ to 0 \quad \algnote{\# cache priority ($i \in [L], j \in [N]$) } \\
Set $\mathbf{H}_{i}(C_{i})$, $\mathbf{P}_{i}(a)$ and $\mathbf{PH}_{i} (a, C_{i})$ to 0 \quad \algnote{\# $ a, C_{i} \in [N]$} \\ \label{alg:init_end}
Set $V_{i}$ to $ V_{e} / L$, and $\theta_{i}$ to 0.5 \quad \algnote{\# uniform allocation} \\ \label{alg:init_strategy}
\While {True}{
    Process a token, and set $q \leftarrow q + 1$ \\ \label{alg:system_variables_update_begin}
    Update $\mathbb{P}_{i, j}$, $\mathbf{H}_{i}(C_{i})$, $\mathbf{P}_{i}(a)$ and $\mathbf{PH}_{i} (a, C_{i})$ \\ \label{alg:system_variables_update_end}
    \If {($q \mod \tau == 0$) \textbf{and} (device is idle)} { \label{alg:cache_update_begin}
        $\{ V_{i} \}_{i \in [L]}, \{\theta_{i}^{*}\}_{i \in [L]} \! \leftarrow \!$ \textbf{VramAllocation} ($V_{e}, \zeta$)  \\ 
        Update the expert cache for each layer \label{alg:cache_update_end}
    }
}

\textbf{\textsc{Function} VramAllocation ($V_{e}, \zeta$)}: \\ \label{alg:function1_begin}

\While{True}{
$ \{\mathcal{T}_{1}^{i}\}_{i \in [L]}, \{\theta_{i}^{*}\}_{i \in [L]}  \leftarrow$ \textbf{ExpertSplit} ($\{V_{i}\}_{i \in [L]}$) \\ \label{alg:t1}
$ \{\mathcal{T}_{2}^{i}\}_{i \in [L]}, \underline{\hspace{2.5mm}} \leftarrow$ \textbf{ExpertSplit} ($\{ V_{i} + \zeta V_{e}\}_{i \in [L]}$) \\ \label{alg:t2}
$ \{\mathcal{T}_{3}^{i}\}_{i \in [L]}, \underline{\hspace{2.5mm}} \leftarrow$ \textbf{ExpertSplit} ($\{ V_{i} - \zeta V_{e}\}_{i \in [L]}$) \\ \label{alg:t3}
$i1 \leftarrow \arg\max_{i \in [L]} ( \mathcal{T}_{1}^{i} - \mathcal{T}_{2}^{i} ) $ \\ \label{alg:i1}
$i2 \leftarrow \arg\min_{i \in [L], i \neq i1} ( \mathcal{T}_{3}^{i} - \mathcal{T}_{1}^{i} )$ \\ \label{alg:i2}
Set $V_{i1} \leftarrow V_{i1} + \zeta V_{e}$, $V_{i2} \leftarrow V_{i2} - \zeta V_{e}$ \\ \label{alg:update_VRAM_budget}
$ \{\mathcal{T}_{4}^{i}\}_{i \in [L]}, \underline{\hspace{2.5mm}} \leftarrow$ \textbf{ExpertSplit} ($\{V_{i}\}_{i \in [L]}$) \\
\If{$\sum_{i =1}^{L} ( \mathcal{T}_{4}^{i} - \mathcal{T}_{1}^{i}) \geq 0$} { \label{alg:check_begin}
    Roll back the opeartion of Line \ref{alg:update_VRAM_budget} \\
    \textbf{Return} $\{ V_{i} \}_{i \in [L]}$, $ \{\theta_{i}^{*}\}_{i \in [L]}$ \\ \label{alg:function1_end}
}
}

\textbf{\textsc{Function} ExpertSplit ($\{V_{i}\}_{i \in [L]}$):}  \\ \label{alg:function2_begin}
Compute $T_{wind}^{1} \leftarrow T_{comp}^{head} + T_{comp}^{att}$ \\ \label{alg:first_layer_prefetch_window}
\For {each layer $i \in [L]$} { \label{alg:for_each_layer}
    Solve the sub-problem in Eq. \eqref{subproblem_formultation} to obtain $\theta_{i}^{*}$ \\ \label{alg:subproblem_solving}
    Calculate $\mathcal{T}^{i}$ and $T_{wind}^{i + 1}$ by Eqs. \eqref{miss_window}-\eqref{prefetch_window} \\ \label{alg:byproducts}
}
\textbf{Return} $ \{\mathcal{T}^{i}\}_{i \in [L]}$, $ \{\theta_{i}^{*}\}_{i \in [L]}$  \label{alg:function2_end}

\end{algorithm}

Based on this sub-problem, we then design a fixed-point iterative algorithm to allocate VRAM budget across different layers, which is formally described in Algorithm \ref{alg}.
At the beginning, the developer inputs latency metrics and VRAM budget into the algorithm and pre-defines two hyper-parameters, including cache update frequency $\tau$, and allocation granularity $\zeta$.
Prior to inference, \method initializes the system variables as 0 (Lines \ref{alg:init_begin}-\ref{alg:init_end}), including tokens counter $q$, each expert $E_{i, j}$'s cache priority $\mathbb{P}_{i, j}$, as well as statistical probabilities $\mathbf{H}_{i}(C_{i})$, $\mathbf{P}_{i}(a)$ and $\mathbf{PH}_{i} (a, C_{i})$ ($\forall i \in [N], \forall j, a, C_{i} \in [N]$).
These variables are updated during each inference pass (Lines \ref{alg:system_variables_update_begin}-\ref{alg:system_variables_update_end}).
Initially, \method adopts a uniform VRAM allocation strategy and an expert split ratio of 0.5 (Line \ref{alg:init_strategy}), with the cached experts selected randomly.
Every $\tau$ processed tokens, \method executes the cache configuration algorithm using the latest state variables (Lines \ref{alg:cache_update_begin}-\ref{alg:cache_update_end}).
It is worth noting that the cache update will be performed only when the device is idle, ensuring that ongoing inference is not affected by the algorithm execution.
After obtaining the algorithm output (\ie, VRAM budget $V_{i}$ and split ratio $\theta_{i}$ for each layer $i$), \method proceeds to configure the expert cache.
Specifically, for each layer $i$, \method ranks all $N$ experts by their cache priorities, and vertically splits the top-$C_{i}$ experts, where $C_{i} = V_{i} / \theta_{i}$.
The top segments of these experts are then cached in VRAM, while their bottom segments, along with all other experts, are offloaded to the CPU RAM.

Next, we present a detailed description of our optimization algorithm.
The algorithm proceeds in an iterative manner (Lines \ref{alg:function1_begin}-\ref{alg:function1_end}), and is built on a fundamental subroutine (Lines \ref{alg:function2_begin}-\ref{alg:function2_end}), which accepts the per-layer VRAM budget as the input.
Concretely, given the constant prefetch window $T_{wind}^{1}$ (Line \ref{alg:first_layer_prefetch_window}) of the first layer, the subroutine sequentially solves the sub-problem for each layer to determine its optimal expert split ratio $\theta_{i}^{*}$ (Lines \ref{alg:for_each_layer}-\ref{alg:subproblem_solving}).
After the determination of each layer $i$, the prefetch window $T_{wind}^{i + 1}$ can be derived to support the sub-problem solving for the next layer.
At the same time, we can also obtain the exposed loading latency $\mathcal{T}^{i}$ of each layer $i$ as byproducts (Line \ref{alg:byproducts}).
Notably, for the calculation of Eqs. \eqref{miss_window}-\eqref{prefetch_window}, the terms $(\alpha_{i} + \beta_{i} \theta_{i})$ and $\beta_{i} (1 - \theta_{i}) + \gamma_{i}$ are approximated by $m_{i}$ and $(K - m_{i})$, respectively.
In each iteration, \method first invokes the subroutine with the current VRAM allocation $\{ V_{i} \}_{i \in [L]}$, to derive the optimal expert split ratios $ \{\theta_{i}^{*}\}_{i \in [L]}$ and the exposed loading latency $\{\mathcal{T}_{1}^{i}\}_{i \in [L]}$ for each layer $i$ (Line \ref{alg:t1}).
Subsequently, \method adjusts the VRAM budget of each layer by adding and subtracting a fixed amount $\zeta V_{e}$, respectively, and calls the subroutines in both cases to derive the per-layer exposed loading latency $\{\mathcal{T}_{2}^{i}\}_{i \in [L]}$ and $\{\mathcal{T}_{3}^{i}\}_{i \in [L]}$ (Lines \ref{alg:t2}-\ref{alg:t3}).
The allocation granularity $\zeta$ lies in the range $(0, 1)$, which is set to 0.01 by default in our experiments.
At this stage, we can use the difference $(\mathcal{T}_{1}^{i} - \mathcal{T}_{2}^{i})$ to estimate the profit of increasing budget for layer $i$, in terms of reduced exposed loading latency.
Similarly, $(\mathcal{T}_{3}^{i} - \mathcal{T}_{1}^{i})$ estimates the increase in exposed loading (\ie, the cost) of reducing budget for layer $i$.
Consequently, the layer with the highest profit and the layer with the smallest cost can be derived (Lines \ref{alg:i1}-\ref{alg:i2}), denoted as $i1$ and $i2$, respectively.
\method then adjusts the VRAM allocation by transferring a budget of $\zeta V_{e}$ from layer $i2$ to layer $i1$ (Line \ref{alg:update_VRAM_budget}).
Next, \method calls the subroutine again to compute the exposed loading latency $ \{\mathcal{T}_{4}^{i}\}_{i \in [L]}$ after the budget adjustment.
Finally, \method checks whether the total profits is positive.
If so, the algorithm continues to the next iteration.
 Otherwise, the algorithm reverts the adjustment of this iteration and returns the current solution (Lines \ref{alg:check_begin}-\ref{alg:function1_end}).

%% file: content/evaluation.tex
\subsection{Experimental Setup} \label{sec:setup}

\textbf{Implementation.}
To evaluate the performance of \method, we utilize Python to implement a customized MoE inference system with expert offloading over an AMAX deep learning workstation, which is equipped with an Intel(R) Xeon(R) Platinum 8358P CPU, 8 NVIDIA RTX A6000 GPUs (48GB), and 512 GB RAM \cite{liu2023finch}.
The data transmission from CPU to GPU is supported by PCIe 4.0x16, with a theoretical maximum bandwidth of 32GB/s.
On the software side, our system is built upon the PyTorch deep learning framework and the HuggingFace Transformers library, which serve as the backbone for model construction and inference execution.
To support the specific requirements of \method, we modified the source code and integrates additional functionalities such as cache item management, cache table lookup, and dynamic expert loading into the model implementation.

\textbf{Models and Datasets.}
As summarized in Table~\ref{table:models}, we adopt two widely used open-source MoE models, \ie, Qwen1.5-MoE \cite{qwen2} and Mixtral-8x7B \cite{jiang2024mixtral}, in our experiments, where Qwen1.5-MoE is evaluated using 16-bit precision.
To accommodate limited computation power and PCIe bandwidth, we adopt quantized versions of Mixtral-8x7B using the half-quadratic quantization (HQQ) technique \cite{badrihalf}, where the attention modules are quantized to 4-bit and the MoE modules to 2-bit.
Additionally, we use the Massive Multitask Language Understanding (MMLU) \cite{hendrycks2020measuring} dataset as a comprehensive benchmark to test \method’s generality across various tasks.
MMLU contains 15,908 questions in total, covering 57 distinct tasks spanning diverse domains, such as STEM, humanities, social sciences, and so on.

\begin{table}[t] 
\centering 
\caption{The MoE models employed in our experiments. \#E, \#P and \#L denote the number of experts per layer, parameters, and layers, respectively. Act. represents the number of parameters or experts activated for each token.}
\vspace{-3mm}
\begin{tabular}{c|cc|cc|c} 
\toprule 
\multirow{2}{*}{Model}  & \multicolumn{2}{c|}{\#E} & \multicolumn{2}{c|}{\#P} & \multirow{2}{*}{\#L} \\  
& Total & Act. & Total & Act. &  \\ 
\midrule 
Qwen1.5-MoE \cite{qwen2}  & 60 & 4 & 14.3B & 2.7B & 24 \\ 
\midrule 
Mixtral-8x7B \cite{jiang2024mixtral}   & 8 & 2 & 46.7B & 12.9B & 32 \\ \bottomrule 
\end{tabular} \label{table:models}
\vspace{-2mm}
\end{table}

\textbf{Baselines.}
To demonstrate the advantages of \method, we compare its performance against four advanced baselines:
\begin{itemize}
    \item \PreGatedMoE \cite{hwang2024pre} trains a pre-gated function for each layer to predict the activated experts for the next layer.
    The predicted experts are prefetched into VRAM in parallel with the computation of the current layer.
    \item \MixtralOffloading \cite{eliseev2023fast} prefetches the experts for the next layer based on the intermediate value of the current layer, and caches a subset of experts via the LRU policy with a fixed cache size for all layers.
    \item \AdapMoE \cite{zhong2024adapmoe} follows \MixtralOffloading to integrate LRU caching and speculative prefetching. 
    Furthermore, it employs an algorithm based on dynamic programming to adjust the cache size for each layer.
    \item  \MoEInfinity \cite{xue2024moe} improves expert prediction accuracy compared to \AdapMoE and \PreGatedMoE by tracing the group activation pattern across multiple layers.
    It employs the LFU policy for expert caching.
\end{itemize}

\textbf{Metric.} 
We quantify the inference performance of \method and baselines using the following metrics.
\begin{itemize}
    \item \textbf{Time-To-First-Token (TTFT)} measures the delay between the arrival of a request and the generation of the first token during the prefill phase, reflecting the system's initial responsiveness.
    \item \textbf{Time-Per-Output-Token (TPOT)} quantifies the interval between the generation of consecutive output tokens during the decode phase, which impacts the overall perceived fluidity of the response.
    \item \textbf{VRAM footprint} is defined as the peak GPU memory usage during the model inference, indicating the memory efficiency and the GPU resource requirements of different approaches.
\end{itemize}

\textbf{Experimental Parameters.} 
For the LCP cache policy, we set the observation window $\omega$ and the balance control parameter $\rho$ as 128 and 0.25 by default, respectively.
For the cache configuration algorithm, the allocation granularity $\zeta$ is set to 0.01.
The algorithm is executed to update the cache configuration every time the system processes 5,000 tokens (\ie, $\tau$ = 5,000).
Moreover, to ensure fair comparison, we set a unified VRAM budget $V_{e}$ for \method and all baselines.
By default, $V_{e}$ is set to the size of 240 complete experts for Qwen1.5-MoE, and 96 complete experts for Mixtral-8x7B.
The buffer size $U_{e}$ is set to the size of $K$ complete experts for both two models.
To ensure statistical significance and mitigate potential outliers, all experimental results reported in this paper represent the average of five independent runs. 

\begin{figure*}[t]
    \centering
    
    \begin{subfigure}
        \centering
        \includegraphics[width=0.65\linewidth]{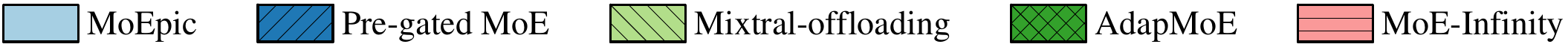}
    \end{subfigure}

    \setcounter{subfigure}{0}
    \subfigure[TTFT with Qwen1.5-MoE.]{
		\includegraphics[width=1.65in]{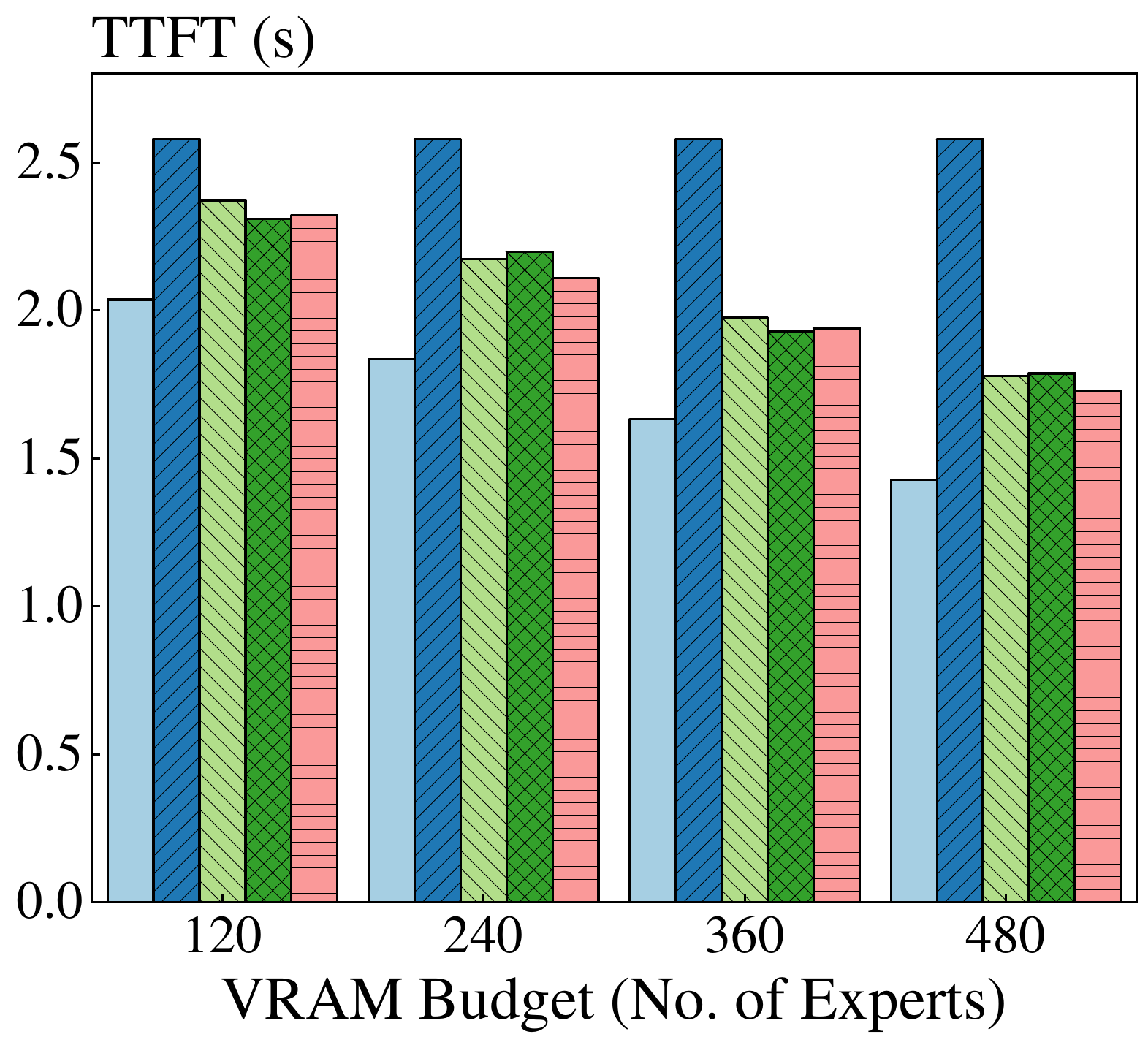}\label{fig:qwen_ttft}
        }
	\subfigure[TTFT with Mixtral-8x7B.]{
		\includegraphics[width=1.65in]{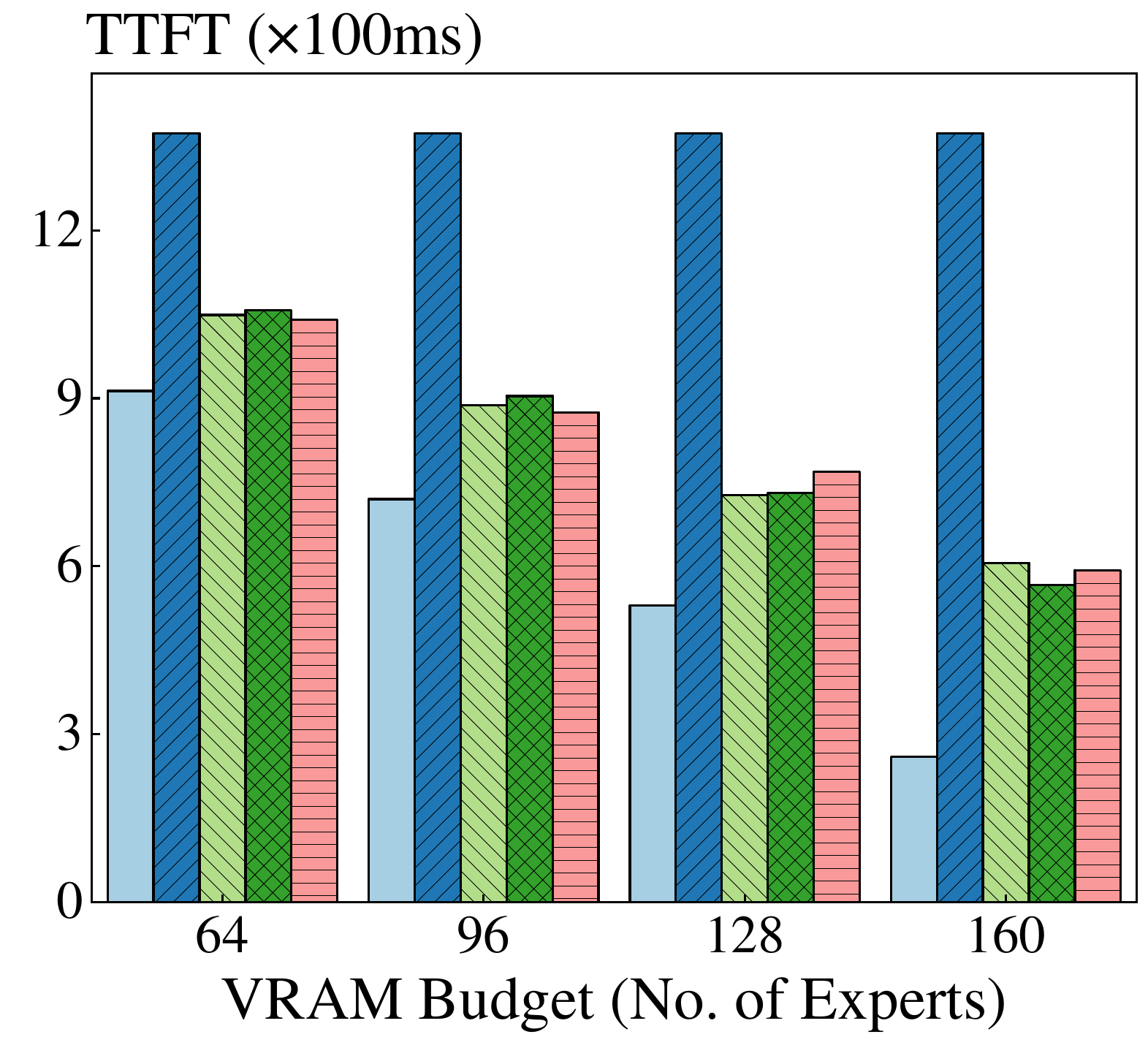}\label{fig:mixtral}
	}
        \subfigure[TPOT with Qwen1.5-MoE.]{
		\includegraphics[width=1.65in]{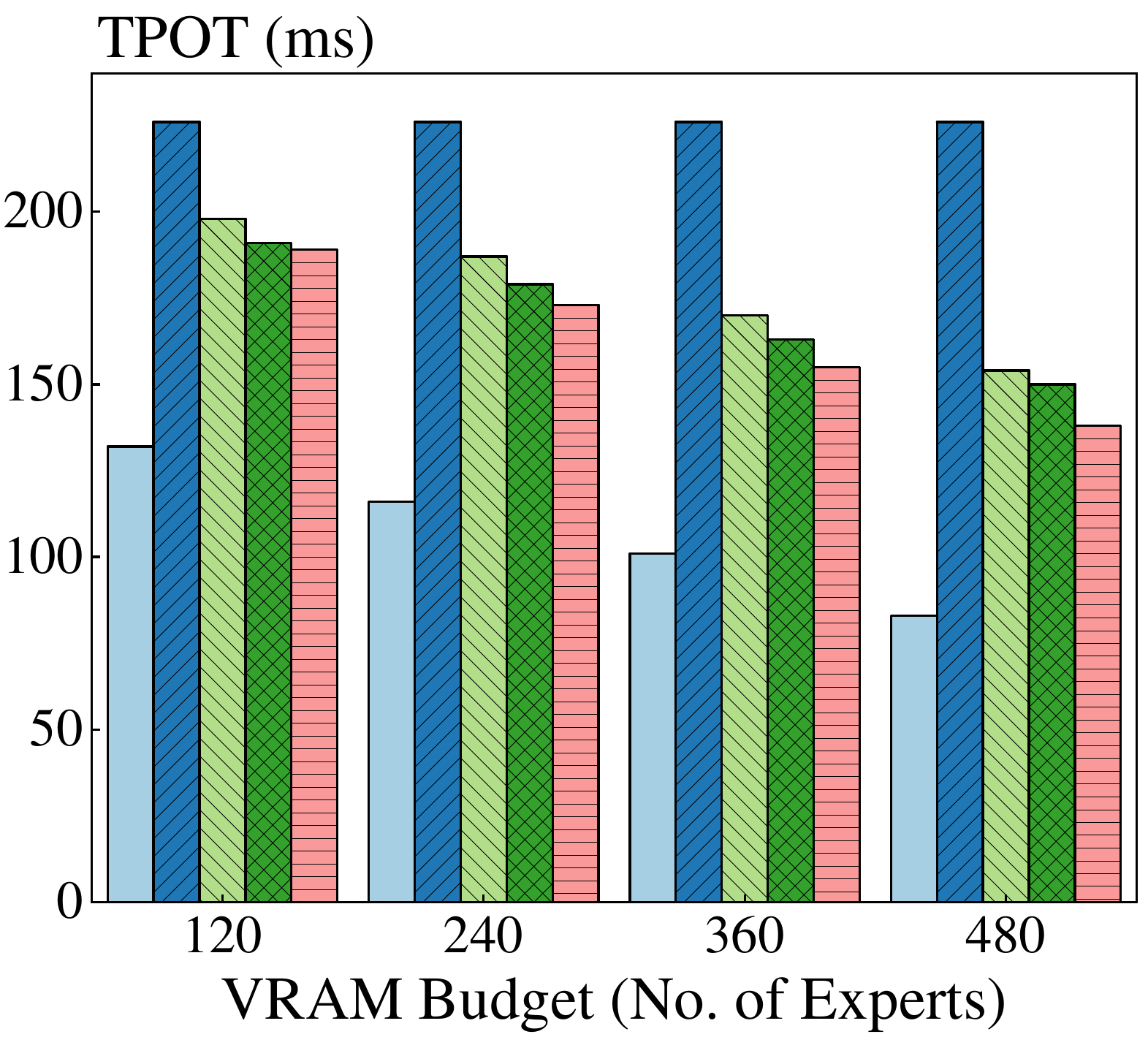}\label{fig:qwen_tpot}
	}
        \subfigure[TPOT with Mixtral-8x7B.]{
		\includegraphics[width=1.65in]{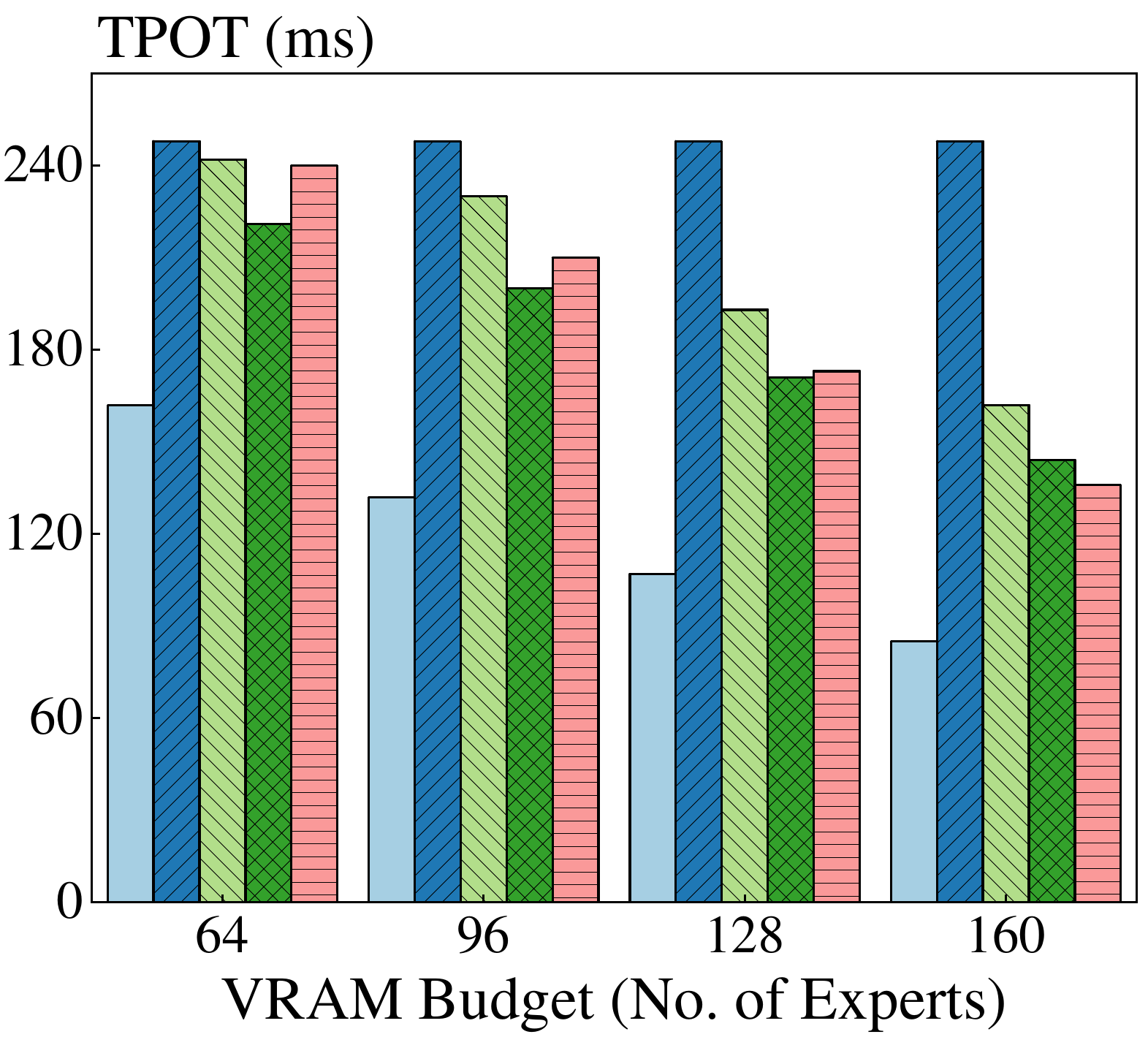}\label{fig:mixtral_tpot}
	}
    \vspace{-3mm}
    \caption{Inference latency (\ie, TTFT and TPOT) of different approaches with Qwen1.5-MoE and Mixtral-8x7B.}
    \label{fig:inference_latency}
\end{figure*}

\subsection{Inference Latency}

In this section, we test the inference latency (\ie, TTFT and TPOT) of \method and baselines under different total VRAM budgets, where the results are shown in Figure \ref{fig:inference_latency}.
It can be observed that the inference latency of all other methods, except for \PreGatedMoE, decreases as the VRAM budget increases.
Moreover, \method consistently achieves the fastest inference speed in both the prefill and decode stages across different settings.
Since there are no cached experts in \PreGatedMoE, variations in the VRAM budget do not affect its performance.
Accordingly, without expert caching, \PreGatedMoE also shows the highest TTFT and TPOT among all methods.
For the prefill stage, experts in each layer are often fully activated, leading to prolonged loading latency.
The \MixtralOffloading, \AdapMoE, and \MoEInfinity baselines achieve comparable performance and surpass \PreGatedMoE, since they only incur the cost of loading non-cached experts.
\method further overlaps the computation of cached experts with the loading of missing experts, achieving the best performance among all approaches.
For instance, by Figure \ref{fig:qwen_ttft}, the TTFT of \method is 1.43s with a VRAM budget of 480 experts in Qwen1.5-MoE, while that of \PreGatedMoE, \MixtralOffloading, \AdapMoE, and \MoEInfinity is 2.58s, 1.77s, 1.78s, and 1.73s, respectively.
In other words, \method can accelerate the prefill speed by about 1.21$\times$$\thicksim$1.80$\times$ compared to the baselines.

As for the decode stage, \MixtralOffloading demonstrates inferior performance compared to \AdapMoE, primarily because it fails to account for the varying cache size requirements across different layers.
By leveraging group activation pattern to improve the prediction accuracy, \MoEInfinity sometimes outperforms \AdapMoE.
For \method, the expert splitting mechanism enables better transfer-computation overlap, and the LCP caching policy achieves significantly higher cache hit rates than LRU and LFU.
Thus, \method also achieves the lowest TPOT among all baselines.
For example, given a VRAM budget of 160 experts in Mixtral-8x7B, \method presents a TPOT of 85ms, while \PreGatedMoE, \MixtralOffloading, \AdapMoE, and \MoEInfinity separately show a TPOT of 248ms, 162ms, 144ms, and 136ms.
As a result, \method can mitigate the TPOT by about 37.51\%$\thicksim$65.73\% compared to the baselines.
These results demonstrate the advantages of \method in inference latency under the setting of expert offloading.

\begin{figure}[t]
    \centering
    
    \begin{subfigure}
        \centering
        \includegraphics[width=0.9\linewidth]{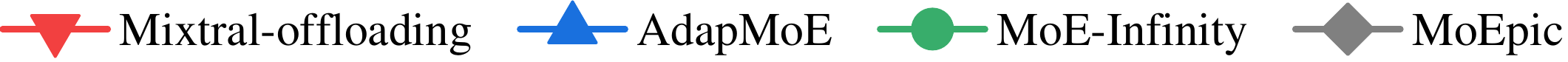}
    \end{subfigure}

    \setcounter{subfigure}{0}
    \subfigure[Footprint with Qwen1.5-MoE.]{
		\includegraphics[width=1.5in]{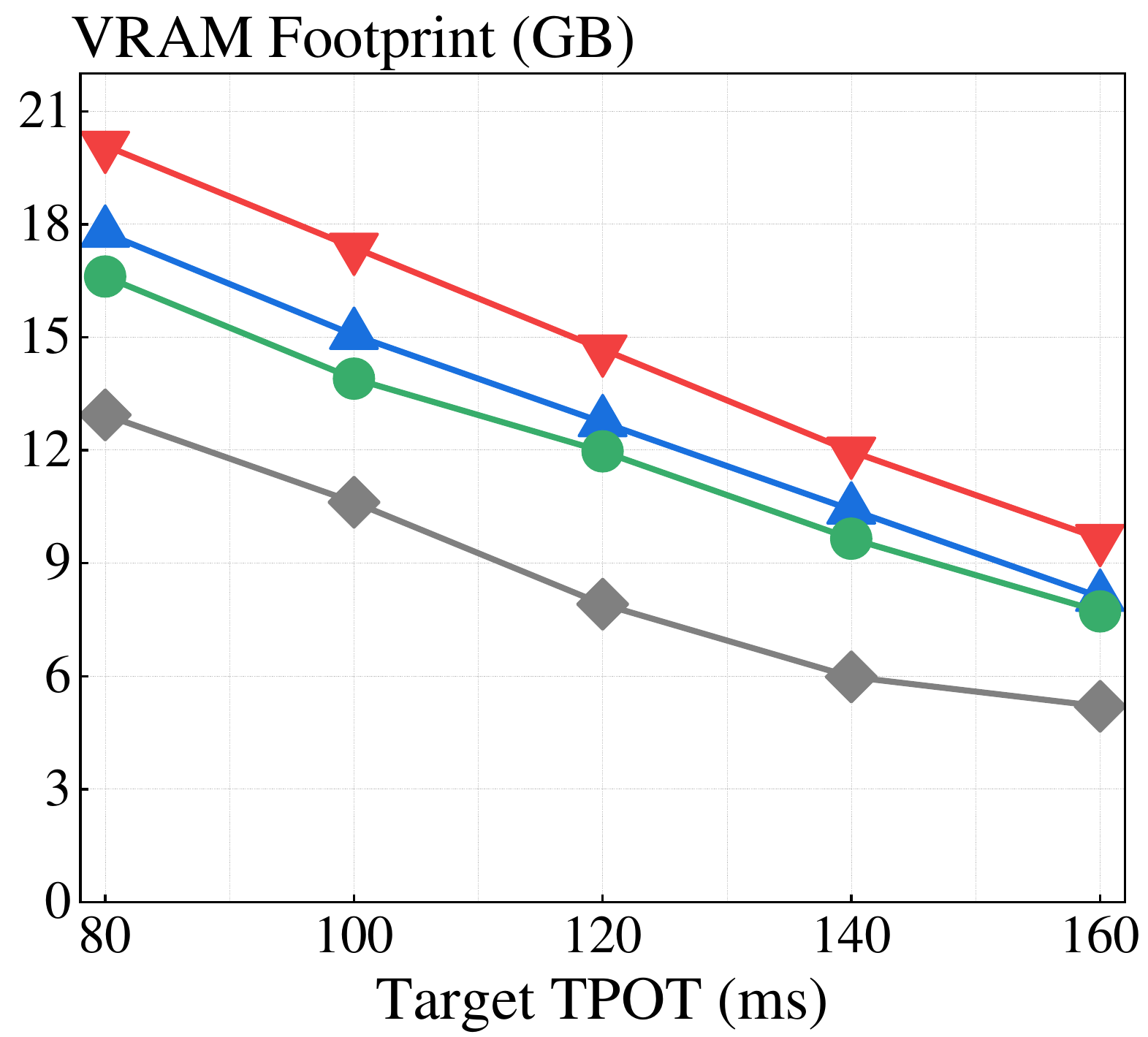}\label{fig:qwen_memory}
        }
	\subfigure[Footprint with Mixtral-8x7B.]{
		\includegraphics[width=1.5in]{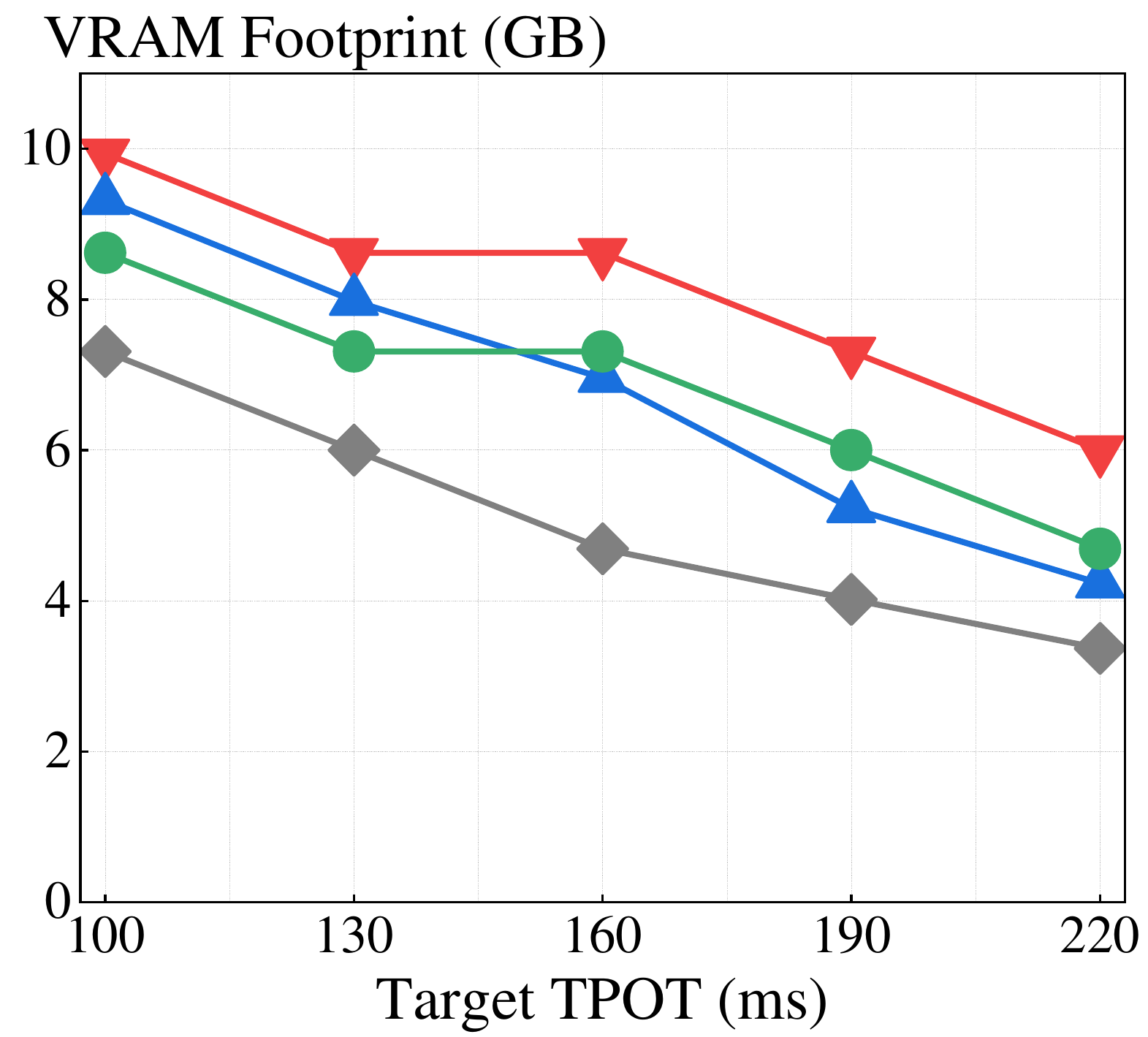}\label{fig:mixtral_memory}
	}
    \vspace{-3mm}
    \caption{VRAM footprint of different approaches under various target TPOT with Qwen1.5-MoE and Mixtral-8x7B.}
    \label{fig:memory_footprint}
    \vspace{-3mm}
\end{figure}

\subsection{Memory Efficiency}

In this section, we establish a common target TPOT across all approaches and evaluate the minimum VRAM footprint each approach requires to meet this target.
In particular, the target TPOT (ms) values are configured as 80, 100, 120, 140, 160 for Qwen1.5-MoE, and 100, 130, 160, 190, 220 for Mixtral-8x7B.
Notably, since \PreGatedMoE does not leverage available VRAM to cache experts, it fails to achieve the target TPOT and is therefore excluded from this set of experiments.
The results are illustrated in Figure \ref{fig:memory_footprint}, indicating that \method delivers the highest memory efficiency among the four approaches.
For instance, by Figure \ref{fig:qwen_memory}, \method requires a VRAM footprint of 5.98GB to reach the target TTFT of 140ms with Qwen1.5-MoE, while \MixtralOffloading, \AdapMoE, and \MoEInfinity need 11.97GB, 10.42GB, and 9.56GB, respectively.
That is, for the same target TPOT, \method can reduce the VRAM footprint by up to 37.45\%$\thicksim$50.04\% compared to the three baselines.
Notably, when the entire Qwen1.5-MoE model is fully placed in VRAM (\ie, GPU-Rich setting), the TPOT is 57ms with a VRAM footprint of 26.67GB.
By contrast, \method requires only 12.38 GB of VRAM to achieve the target TPOT of 60 ms.
Thus, \method can save about half of the GPU cost for the MoE inference while incurring only a 5.26\% increase in inference latency.

\begin{table}[t] 
\centering 
\caption{Ablation study of \method with Mixtral-8x7B.}
\vspace{-2mm}
\begin{tabular}{l|cc|cc} 
\toprule 
\multirow{2}{*}{Approaches}  & \multicolumn{2}{c|}{Prefill} & \multicolumn{2}{c}{Decode} \\  
& TTFT & Speed$\downarrow$  & TPOT &  Speed$\downarrow$  \\ 
\midrule 
\method  & 720ms & -- & 132ms & --  \\ 
\midrule 
w/o SP   & 753ms & 1.05$\times$ & 174ms &  1.32$\times$  \\ 
\midrule 
w/o LCP  & 782ms & 1.09$\times$ & 157ms & 1.19$\times$ \\ 
\midrule 
w/o CCA  & 864ms & 1.21$\times$ & 193ms & 1.46$\times$ \\ 
\bottomrule 
\end{tabular} \label{table:ablation}
\vspace{-2mm}
\end{table}

\subsection{Ablation Study}

The \method system comprises three key components: speculative prefetcher (SP), priority-based cache policy (LCP), and cache configuration algorithm (CCA).
In this section, we conduct a set of experiments with Mixtral-8x7B to evaluate the effect of these three components.
For \method without SP, experts are prefetched randomly within each layer's computation window.
In \method without LCP, the LCP policy is replaced by a random (RND) cache replacement strategy.
As for \method without CCA, we employ uniform VRAM budget allocation across layers and fix the expert split ratio at 0.5.
The results in Table \ref{table:ablation} demonstrate that CCA plays a crucial role in both the prefill and decode stages.
Specifically, without CCA, the TTFT and TPOT of \method are prolonged by about 1.21$\times$ and 1.46$\times$, respectively.
Although the SP and LCP components positively contribute to \method’s performance, their impact is relatively modest, especially in the prefill stage.
These results indicate the necessity and importance of all three components, while CCA plays the most
critical role in \method’s performance.
\vspace{-2mm}
\subsection{Sensitivity Analysis}

\begin{figure}[t]
    \centering

    \setcounter{subfigure}{0}
    \subfigure[Impact of $\rho$ and $\omega$.]{
		\includegraphics[width=1.5in]{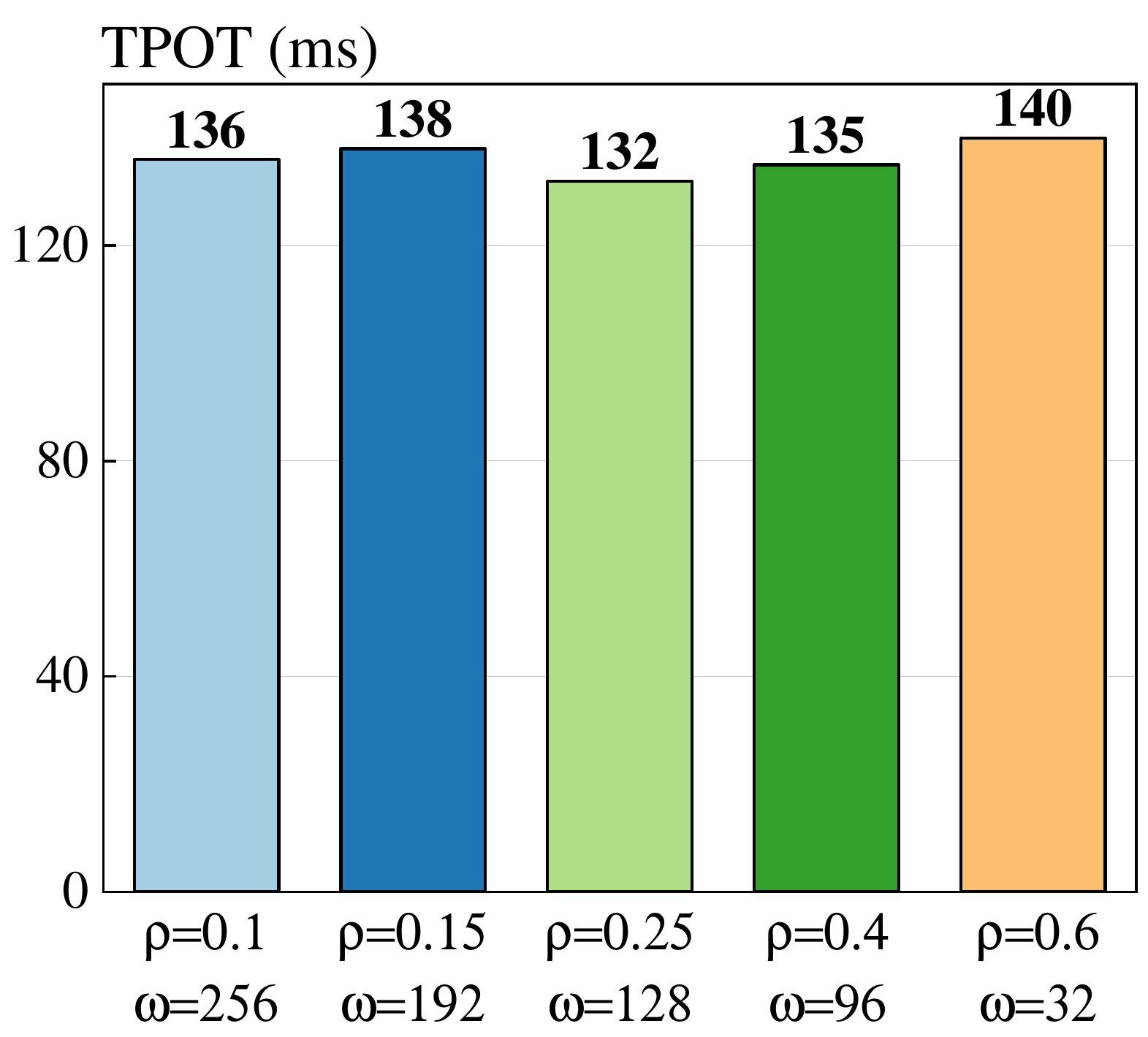}\label{fig:rho_omega}
        }
	\subfigure[Impact of $\zeta$,]{
		\includegraphics[width=1.5in]{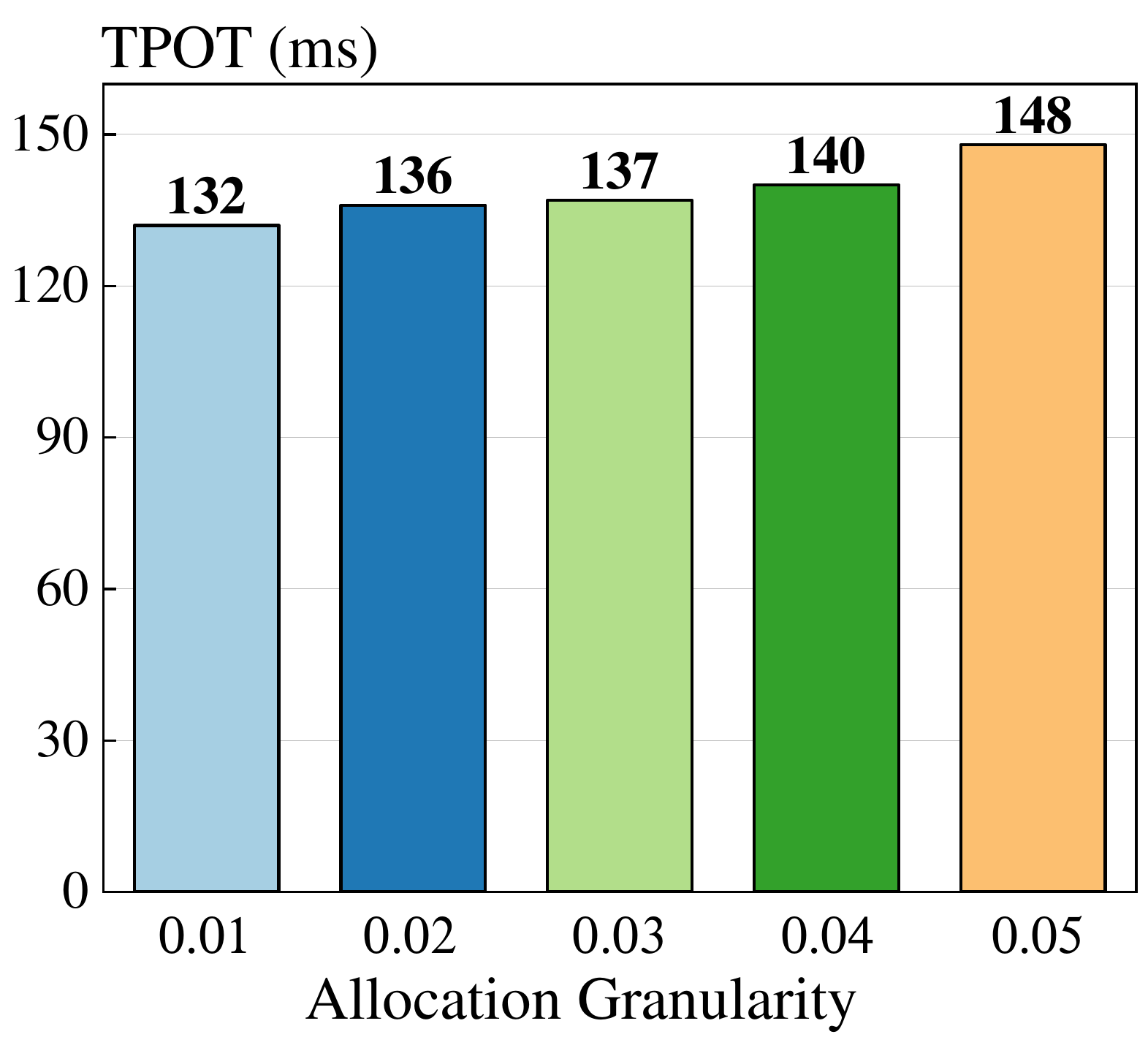}\label{fig:zeta}
	}
    \vspace{-3mm}
    \caption{Impact of different hyper-parameters settings on \method's performance with Mixtral-7x8B.}
    \label{fig:memory_footprint}
    \vspace{-3mm}
\end{figure}

In this section, we conduct a set of experiments using Mixtral-8x7B to measure \method's performance (\ie, TPOT) with different settings of hyper-parameters, including the balance control parameter $\rho$ and observation window $\omega$ of the LCP policy, as well as the allocation granularity $\zeta$ of the cache configuration algorithm.
Firstly, different settings of $\rho$ and $\omega$ only affect the cache hit rate slightly, without causing significant impacts to the inference latency.
Specifically, we present the TPOT of \method under five different settings of $\rho$ and $\omega$ in Figure \ref{fig:rho_omega}, where the latency variation is at most 8ms.
Secondly, a smaller granularity $\zeta$ allows for more flexible VRAM budget allocation, enabling better inference performance.
As shown in Figure \ref{fig:zeta}, \method's TPOT increases as the granularity $\zeta$ becomes larger.
For instance, when $\zeta$ is set to 0.05, the latency is prolonged by about 12.12\% compared to the default setting (\ie, $\zeta = 0.01$).
However, while a finer allocation granularity may yield performance gains, it also increases the number of iterations required for the algorithm to converge. 
Overall, the performance of \method exhibits minimal sensitivity to variations in hyper-parameter settings, underscoring its strong robustness.

%% file: content/works.tex
Although LLMs demonstrate remarkable capabilities, their substantial VRAM requirements for inference lead to prohibitively high hardware costs for local deployment, particularly for MoE-based LLMs.
In recent years, many studies have focused on reducing the VRAM footprint of MoE inference, which can be broadly divided into the following two classes:

\textbf{Expert Compression.}
These approaches aim to slim the expert parameters using techniques such as pruning \cite{limerge, lu2024not}, quantization \cite{huang2024mixture}, distillation \cite{yuan2023compressed}, and tensor decomposition \cite{yang2024moe}.
For instance, \textsf{NAEE} \cite{lu2024not} adopts a proxy dataset to evaluate the contribution of each expert to the inference accuracy, and eliminates the low-contribution experts from each model layer.
\textsf{MC-SMoE} \cite{limerge} divides experts into different groups and merges each group to one expert, thus reducing the number of experts.
\textsf{MC-MoE} \cite{huang2024mixture} estimates each expert's importance based on its activation frequency and router weight, which then guide the assignment of quantization precision for each expert.
\textsf{CMoE} \cite{yuan2023compressed} leverages knowledge distillation to transfer knowledge from the MoE model to a smaller dense model, thereby reducing the overall parameter count.
\textsf{MoE-$I^2$} \cite{yang2024moe} decomposes each dense expert matrix into two small low-rank tensors, and adaptively adjusts the ranks for different experts according to their importance.
However, while these methods can reduce the model size, compression inevitably incurs accuracy degradation.
Moreover, the compressed models typically require additional retraining, introducing extra computational cost.
Therefore, these approaches offer a relatively low cost-effectiveness.

\textbf{Expert Offloading.}
This line of research addresses the VRAM constraints by offloading experts to CPU RAM or SSD and retrieving them on demand during inference.
Nevertheless, the expert loading latency significantly degrades the inference speed.
To tackle this issue,  \PreGatedMoE \cite{hwang2024pre} proposes an expert prefetching architecture, overlapping the GPU computation and expert loading.
As an optimization, some approaches \cite{ren2024enabling, song2024promoe, he2024expertflow, du2024sida} train lightweight proxy networks to predict the required experts (\eg, MLP, LSTM), thereby improving the prefetching accuracy.
\MoEInfinity \cite{xue2024moe} tracks the request-level usage frequency of each expert and prefetches high-priority experts.
\MixtralOffloading \cite{eliseev2023fast} employs a hybrid approach combining speculative expert prefetching with LRU-based expert caching. 
Building upon this, \AdapMoE \cite{zhong2024adapmoe}     introduces adaptive cache allocation across different model layers, enabling better performance.
\textsf{HOBBIT} \cite{tang2024hobbit} quantifies the experts at various bit widths, and prioritizes high-precision experts for caching.
However, these approaches still suffer from the poor cache hit rate and high expert loading latency, which bottleneck the inference speed.
Therefore, we propose \method to overcome the challenges of expert offloading.

%% file: content/conclusion.tex
In this paper, we present an efficient MoE inference system, called \method, under the expert offloading setting.
\method integrates the expert caching and speculative prefetching techniques.
It utilizes a novel expert split mechanism to simultaneously improve the number of cachable experts and reduce the prefetching latency, under the constraint VRAM budget and PCIe bandwidth.
Besides, to fully unleash the potential of \method, we further design a divide-and-conquer algorithm to determine proper VRAM budgets and expert split ratios for different layers.
Besides, a new cache policy, named LCP, is introduced in this paper to enable better cache management.
Finally, we conduct extensive experiments to evaluate the performance of \method, and the results show that \method achieves significant improvements in inference latency and memory efficiency compared to the baselines.